%% file: main.tex
\documentclass[letterpaper]{article} % DO NOT CHANGE THIS
\usepackage[preprint]{aaai2027}  % DO NOT CHANGE THIS
\usepackage[hyphens]{url}  % DO NOT CHANGE THIS
\usepackage{graphicx} % DO NOT CHANGE THIS
\usepackage{natbib}  % DO NOT CHANGE THIS AND DO NOT ADD ANY OPTIONS TO IT
\usepackage{caption} % DO NOT CHANGE THIS AND DO NOT ADD ANY OPTIONS TO IT
\usepackage{algorithm}
\usepackage{algorithmic}
\usepackage{multirow}
\usepackage{newfloat}
\usepackage{listings}
\DeclareCaptionStyle{ruled}{labelfont=normalfont,labelsep=colon,strut=off} % DO NOT CHANGE THIS
\floatstyle{ruled}
\newfloat{listing}{tb}{lst}{}
\floatname{listing}{Listing}

\usepackage{booktabs}

\usepackage{amsmath}
\usepackage{amssymb}

\usepackage{colortbl}

\title{DexMani: Human-Derived Manipulability Guidance for Dexterous Rotation}
\author{
    Xiaoyang Chen$^{1,2,3}$\equalcontrib, Shengcheng Luo$^{2,3}$\equalcontrib, Haoran Guo$^{2}$, Jiaming Jiang$^{2,3}$, \\ 
    Wanlin Li$^{3}$, Ziyuan Jiao$^{3,4}$\corresponding, Chenxi Xiao$^{2}$\corresponding
}
\affiliations{
    \textsuperscript{\rm 1}Shanghai Jiao Tong University \quad \textsuperscript{\rm 2}ShanghaiTech University \\
    \textsuperscript{\rm 3}Beijing Institute for General Artificial Intelligence (BIGAI) \quad  \textsuperscript{\rm 4}Beihang University
}

\begin{document}

\maketitle

\begin{abstract}
\input{Tex/0_abstract}
\end{abstract}


\section{1  Introduction}
\input{Tex/1_intro}

\section{2  Related Work}
\input{Tex/2_rela}

\section{3  Methods}
\input{Tex/3_method}

\begin{table*}[t]
\centering
\begin{tabular}{l*{3}{cc}c}
\toprule
\multirow{2}{*}{\textbf{Method}}
& \multicolumn{2}{c}{\textbf{Unscrew Cap}}
& \multicolumn{2}{c}{\textbf{Rotate Object}}
& \multicolumn{2}{c}{\textbf{Turn Faucet}}
& \multirow{2}{*}{\textbf{Avg. SR}} \\
\cmidrule(lr){2-3}
\cmidrule(lr){4-5}
\cmidrule(lr){6-7}
& \textbf{Seen}
& \textbf{Unseen}
& \textbf{Seen}
& \textbf{Unseen}
& \textbf{Seen}
& \textbf{Unseen}
& \\
\midrule

PPO~\cite{schulman2017proximal}
& $11.8 \pm {\scriptstyle 2.1}$
& $5.0 \pm {\scriptstyle 1.9}$
& $26.5 \pm {\scriptstyle 0.7}$
& $21.5 \pm {\scriptstyle 0.6}$
& $33.2 \pm {\scriptstyle 3.5}$
& $20.5 \pm {\scriptstyle 1.0}$
& 19.8 \\

VT Pretraining
& $29.0 \pm {\scriptstyle 1.5}$
& $12.7 \pm {\scriptstyle 3.1}$
& $42.6 \pm {\scriptstyle 1.4}$
& $34.0 \pm {\scriptstyle 4.6}$
& $62.4 \pm {\scriptstyle 4.7}$
& $53.5 \pm {\scriptstyle 4.8}$
& 39.0 \\

VTM
& $47.0 \pm {\scriptstyle 0.9}$
& $27.8 \pm {\scriptstyle 1.8}$
& $49.5 \pm {\scriptstyle 3.0}$
& $37.9 \pm {\scriptstyle 3.4}$
& $79.0 \pm {\scriptstyle 1.2}$
& $70.1 \pm {\scriptstyle 3.2}$
& 51.9 \\

VTA
& $40.1 \pm {\scriptstyle 0.8}$
& $25.1 \pm {\scriptstyle 1.7}$
& $37.4 \pm {\scriptstyle 5.8}$
& $20.3 \pm {\scriptstyle 1.8}$
& $77.2 \pm {\scriptstyle 2.2}$
& $63.5 \pm {\scriptstyle 7.7}$
& 43.9 \\

VTA-E
& $24.2 \pm {\scriptstyle 2.0}$
& $11.6 \pm {\scriptstyle 3.6}$
& $29.0 \pm {\scriptstyle 6.3}$
& $19.1 \pm {\scriptstyle 3.0}$
& $67.6 \pm {\scriptstyle 1.9}$
& $42.0 \pm {\scriptstyle 5.1}$
& 32.3 \\

\rowcolor[gray]{0.93}
\textbf{DexMani}
& $\mathbf{65.1} \pm {\scriptstyle 2.4}$
& $\mathbf{39.3} \pm {\scriptstyle 2.5}$
& $\mathbf{50.7} \pm {\scriptstyle 2.2}$
& $\mathbf{39.2} \pm {\scriptstyle 3.1}$
& $\mathbf{80.2} \pm {\scriptstyle 2.0}$
& $\mathbf{70.3} \pm {\scriptstyle 2.9}$
& \textbf{57.5} \\

\bottomrule
\end{tabular}
\caption{\textbf{Peformance of generalization across tasks and objects.}
Using a single frozen human-derived prior, DexMani achieves the highest mean SR in all six task--split settings, including Turn Faucet, which is absent from human pretraining.}
\label{tab:task_transfer}
\end{table*}

\begin{table*}[t]
\centering
\begin{tabular}{l*{3}{cc}c}
\toprule
\multirow{2}{*}{\textbf{Method}}
& \multicolumn{2}{c}{\textbf{Shadow Hand}}
& \multicolumn{2}{c}{\textbf{Allegro Hand}}
& \multicolumn{2}{c}{\textbf{XHand}}
& \multirow{2}{*}{\textbf{Avg. SR}} \\
\cmidrule(lr){2-3}
\cmidrule(lr){4-5}
\cmidrule(lr){6-7}
& \textbf{Seen}
& \textbf{Unseen}
& \textbf{Seen}
& \textbf{Unseen}
& \textbf{Seen}
& \textbf{Unseen}
& \\
\midrule

PPO~\cite{schulman2017proximal}
& $20.8 \pm {\scriptstyle 2.6}$
& $6.8 \pm {\scriptstyle 2.6}$
& $11.8 \pm {\scriptstyle 1.6}$
& $5.2 \pm {\scriptstyle 2.4}$
& $32.5 \pm {\scriptstyle 3.3}$
& $17.2 \pm {\scriptstyle 1.6}$
& 15.7 \\

VTM
& $66.4 \pm {\scriptstyle 4.4}$
& $40.9 \pm {\scriptstyle 9.2}$
& $24.3 \pm {\scriptstyle 3.9}$
& $12.5 \pm {\scriptstyle 4.2}$
& $58.0 \pm {\scriptstyle 1.4}$
& $22.7 \pm {\scriptstyle 8.4}$
& 37.5 \\

VTA-E
& $52.7 \pm {\scriptstyle 3.2}$
& $24.5 \pm {\scriptstyle 6.7}$
& $17.6 \pm {\scriptstyle 3.5}$
& $9.8 \pm {\scriptstyle 1.7}$
& $45.2 \pm {\scriptstyle 2.8}$
& $18.2 \pm {\scriptstyle 7.5}$
& 28.0 \\

\rowcolor[gray]{0.93}
\textbf{DexMani}
& $\mathbf{70.0} \pm {\scriptstyle 3.1}$
& $\mathbf{48.2} \pm {\scriptstyle 4.8}$
& $\mathbf{34.6} \pm {\scriptstyle 4.0}$
& $\mathbf{14.8} \pm {\scriptstyle 2.3}$
& $\mathbf{63.4} \pm {\scriptstyle 4.5}$
& $\mathbf{29.5} \pm {\scriptstyle 1.9}$
& \textbf{43.4} \\

\bottomrule
\end{tabular}

\caption{\textbf{Performance of Cross-hand generalization on cap-unscrewing task.}
The same human-derived DexMani prior is reused across the Shadow, Allegro, and XHand platforms, with a separate policy trained in each hand-specific action space.}
\label{tab:hand_transfer}
\end{table*}

\section{4 Experiments}

\input{Tex/4_exp}

\section{5  Conclusion}
\input{Tex/5_conlcusion}

\section{Acknowledgment}
This work was supported by the National Natural Science Foundation of China (Grant No. 52305007), the Natural Science Foundation of Shanghai (Grant No. 25ZR1402370), the Artificial Intelligence Project of the State Key Laboratory of General Artificial Intelligence, BIGAI, Peking University, Beijing, China (Project No. SKLAGI2025OP19), the State Key Laboratory of Mechanical System and Vibration (Grant No. MSV202519) and the MoE Key Laboratory of Intelligent Perception and Human-Machine Collaboration (KLIP-HuMaCo).

\bibliography{aaai2027}

\clearpage
\twocolumn[
\begin{center}
{\LARGE\bfseries Supplementary Material\par}
\vspace{1em}
\end{center}
]

\input{sup_arxiv}

\end{document}

%% file: Tex/0_abstract.tex
Dexterous object rotation is a sequential contact problem: each support, release, and re-contact decision must both produce the desired object motion, and prepare the hand configuration for continued rotation. Existing reinforcement learning methods discover such movement patterns through trial and error on specific robotic hand embodiments, without explicitly accounting for how each contact transition affects the hand’s ability to sustain object rotation in subsequent steps. We introduce \textbf{DexMani}, a framework that transfers human demonstrations as \emph{contact-conditioned manipulability evolution}. This prior captures how successful human contact transitions reshape the object-rotation directions available to the hand. DexMani then learns this manipulability evolution and uses it to guide downstream reinforcement learning, enabling rotation skills to be acquired across robot embodiments with distinct kinematics and active-contact configurations. Across the Shadow Hand, Allegro Hand, and XHand, DexMani achieves the highest success rates in every evaluated setting for both seen and unseen objects. DexMani reaches an average success rate of 57.5\% on LEAP Hand, outperforming other baselines and producing smoother rotatory motions. \begin{links}
    \link{Project site}{https://dexmani.github.io/}
\end{links}

%% file: Tex/1_intro.tex
Dexterous rotation tasks, such as unscrewing a cap, turning a valve, or spinning an object, often require the fingers to repeatedly break and re-establish contact~\cite{morgan2022complex}. Each contact transition must not only advance the object’s rotation but also leave the hand in a new configuration that support continued rotation about the desired axis. We refer to this configuration-dependent capability as \emph{contact-conditioned rotational manipulability}. Sustained rotation therefore depends not only on the immediate rotational progress of each transition, but also on how this manipulability is maintained and reshaped over successive transitions.

Despite its importance for generating rotation motion, the evolution of contact-conditioned rotational manipulability (hereafter referred to simply as \emph{manipulability}) has received little attention in existing learning-based methods. For instance, reinforcement learning methods can discover effective contact transitions through extensive trial and error, but generally do not explicitly consider how manipulability evolves with the hand's contact state~\cite{qi2023general,yang2024anyrotategravityinvariantinhandobject}. Similarly, imitation via demonstration-retargeting approaches transfer human motions into embodiment-specific robot trajectories~\cite{qin_dexmv,li2025maniptrans}, but have not yet incorporated contact-conditioned manipulability evolution as an explicit guidance or conditioning signal.

\begin{figure*}[t]
\centering
\includegraphics[width=\textwidth]{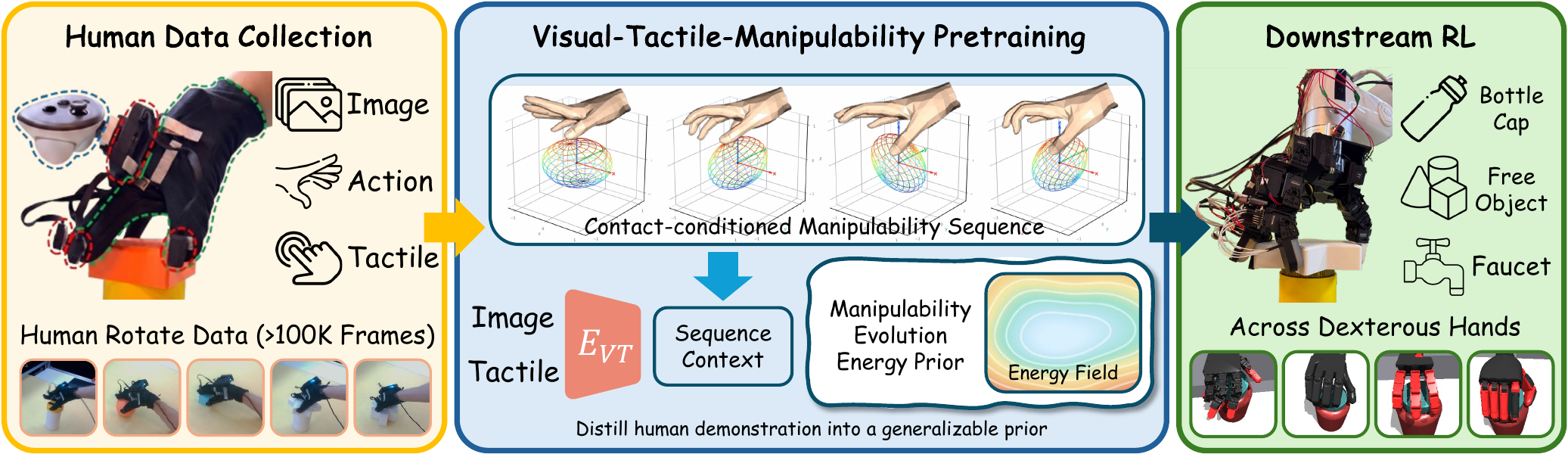}
\caption{\textbf{DexMani overview.} Human visual--tactile rotation demonstrations are used to pretrain an energy-based prior over contact-conditioned manipulability evolution. The frozen prior then guides downstream reinforcement learning across rotation tasks and dexterous hand embodiments.}
\label{fig:teaser}
\end{figure*}

To bridge the gap, we propose \textbf{DexMani}, a framework that distills successful human rotation demonstrations into a prior over the short-horizon evolution of rotational manipulability, and uses it to guide downstream robot learning (Fig.~\ref{fig:teaser}). DexMani first learns an energy model that evaluates changes in manipulability based on patterns observed in human demonstrations. During robot control, the model scores the robot manipulability changes induced by candidate actions, and then generate residual action guidance that favors human-like evolution. Because this guidance is expressed through a manipulability representation that is transferable across embodiments, DexMani can generalize across rotation tasks and dexterous hands.

To evaluate its performance, we conducted comprehensive evaluations of DexMani on three contact-rich rotation tasks: Unscrew Cap, Rotate Object, and Turn Faucet. The results demonstrate strong cross-embodiment and cross-task generalization, with DexMani achieving the highest success rates on both seen and unseen objects in every evaluated setting. It achieves an average success rate of 57.5\% across the three LEAP Hand tasks and 43.4\% for cap unscrewing across the Shadow, Allegro, and XHand embodiments. We additionally demonstrate closed-loop deployment on a physical dexterous hand, highlighting the practical applicability of the proposed approach to real-world robotic systems.

In summary, our main contributions are:
\begin{itemize}
    \item We introduce contact-conditioned manipulability, a human-to-robot representation that captures how contact transitions reshape task-space rotational capabilities.

    \item We develop an energy-based prior that evaluates whether changes in manipulability are consistent with human demonstrations and conducive to sustained rotation.

    \item We propose and evaluate a reinforcement learning framework that transfers this shared prior across dexterous hands with different kinematics.
\end{itemize}

%% file: Tex/2_rela.tex
\subsection{Learning-Based Dexterous In-Hand Rotation}

Learning-based methods generally acquire dexterous rotation skills from robot-generated interaction data, human demonstrations, or a combination of both. For instance, reinforcement learning has been applied to a variety of rotation tasks, including in-hand reorientation~\cite{qi2023general}, tactile-only control~\cite{yin2023rotating}, rapid finger gaiting~\cite{wang2024penspin}, and articulated-object rotation~\cite{yang2024anyrotategravityinvariantinhandobject}. More recent work has explored transferable interaction representations learned from large-scale visual--tactile data~\cite{doi:10.1126/scirobotics.ady2869,huang2026ht,zhang2026humancentrictransferabletactilepretraining}, leveraging privileged simulation information, policy distillation, and tactile perception to enable effective rotation skills. However, these methods must still discover effective contact transitions through either trial-and-error exploration or imitation of human demonstrations. They lack an explicit prior that encodes the desired motion trend throughout the rotation process.

\subsection{Human Demonstration Priors and Motion Transfer}

Recent research has increasingly used human demonstrations to guide robot learning. These approaches often convert demonstrations into hand-specific actions that a robot can reproduce. For example, a commonly used method is to retarget human motions into robot-specific joint commands, or fingertip trajectories~\cite{qin_dexmv,li2025maniptrans}. Recent work has further improved retargeting through contact priors, or by models for correcting retarget errors~\cite{wu2026toporetargetinteractionpreservingretargetingdexterous,pan2026spiderscalablephysicsinformeddexterous}. Beyond retargeting, demonstration-guided reinforcement learning incorporates human-motion priors through policy pretraining, or additional reward objectives~\cite{Rajeswaran-RSS-18,2021-TOG-AMP,lum2025crossinghumanrobotembodimentgap}. However, these representations are often embodiment-specific: differences in hand kinematics and workspace necessitate retargeting or policy relearning for each new hand. In contrast, DexMani transfers an embodiment-agnostic prior over the desired evolution of manipulability, allowing each hand to realize this trend through its own actions and contact transitions.

\subsection{Manipulability and Cross-Embodiment Guidance}

Manipulability characterizes the directions of motion enabled by a hand configuration~\cite{friedman2007task,yokokohji2009dynamic}. Prior work has investigated the geometry of manipulability ellipsoids~\cite{jaquier2021geometry}, transferred desired manipulability profiles across embodiments~\cite{jaquier2020analysistransferhumanmovement,human2robot_manipulability}, and incorporated manipulability objectives into policy learning~\cite{RMT,limanidp}. However, these methods do not explicitly account for how object contacts constrain manipulability during rotation. DexMani addresses this limitation with a contact-conditioned manipulability measure that captures a hand's ability to sustain rotational motion under contact constraints, providing embodiment-agnostic guidance for contact-rich manipulation.

%% file: Tex/3_method.tex
\begin{figure*}[t]
\centering
\includegraphics[width=\textwidth]{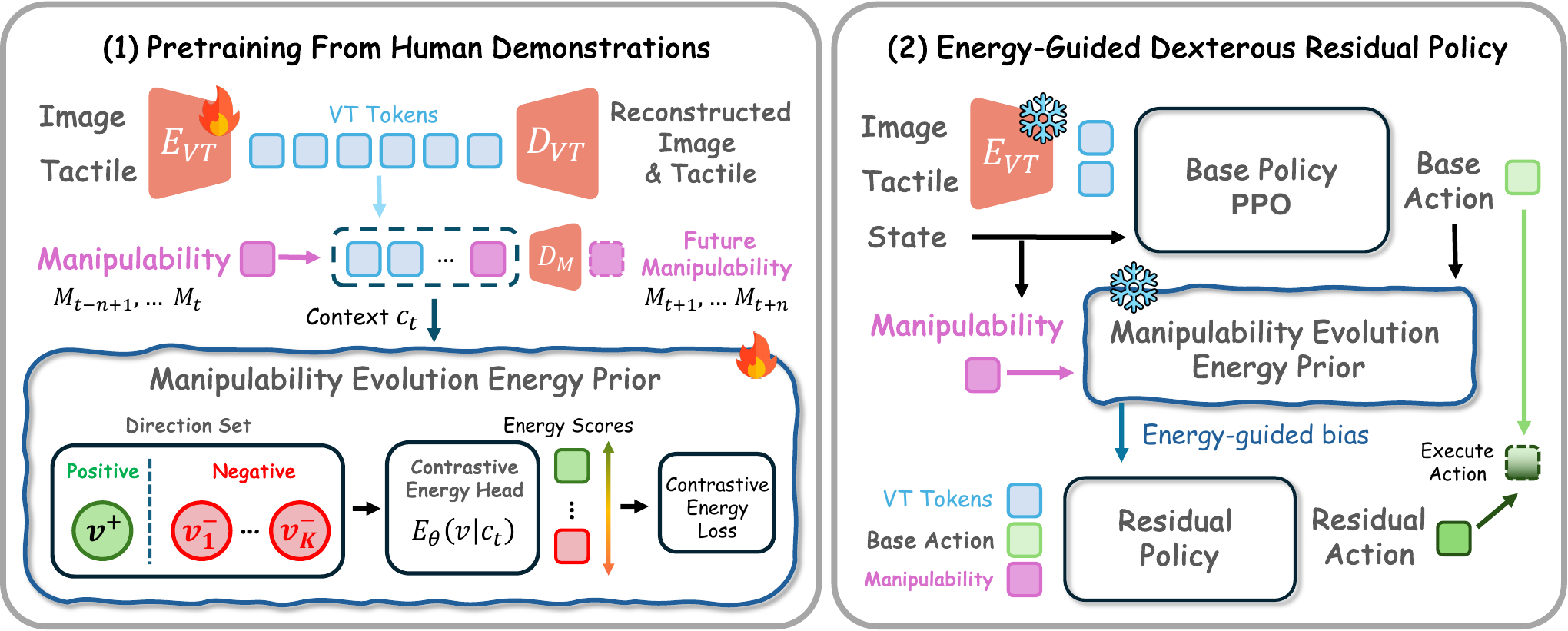}
\caption{\textbf{DexMani pipeline.} DexMani learns a contact-conditioned manipulability-evolution energy prior from human visual--tactile demonstrations (left). During robot reinforcement learning, the frozen prior scores action-induced manipulability changes and provides hand-specific guidance to a residual policy (right).}
\label{fig:pretrain}
\end{figure*}

DexMani aims to transfer a preference over how rotational capability evolves during manipulation. As shown in Fig.~\ref{fig:pretrain}, it first learns such a preference prior from human demonstrations. Then, during downstream reinforcement learning, the learned prior is used to provide residual action guidance, encouraging the policy to follow the learned preference.

\subsection{3.1 Human Demonstration Collection}
\label{sec:method:human_data}

\begin{figure}[t]
\centering
\includegraphics[width=0.47\textwidth]{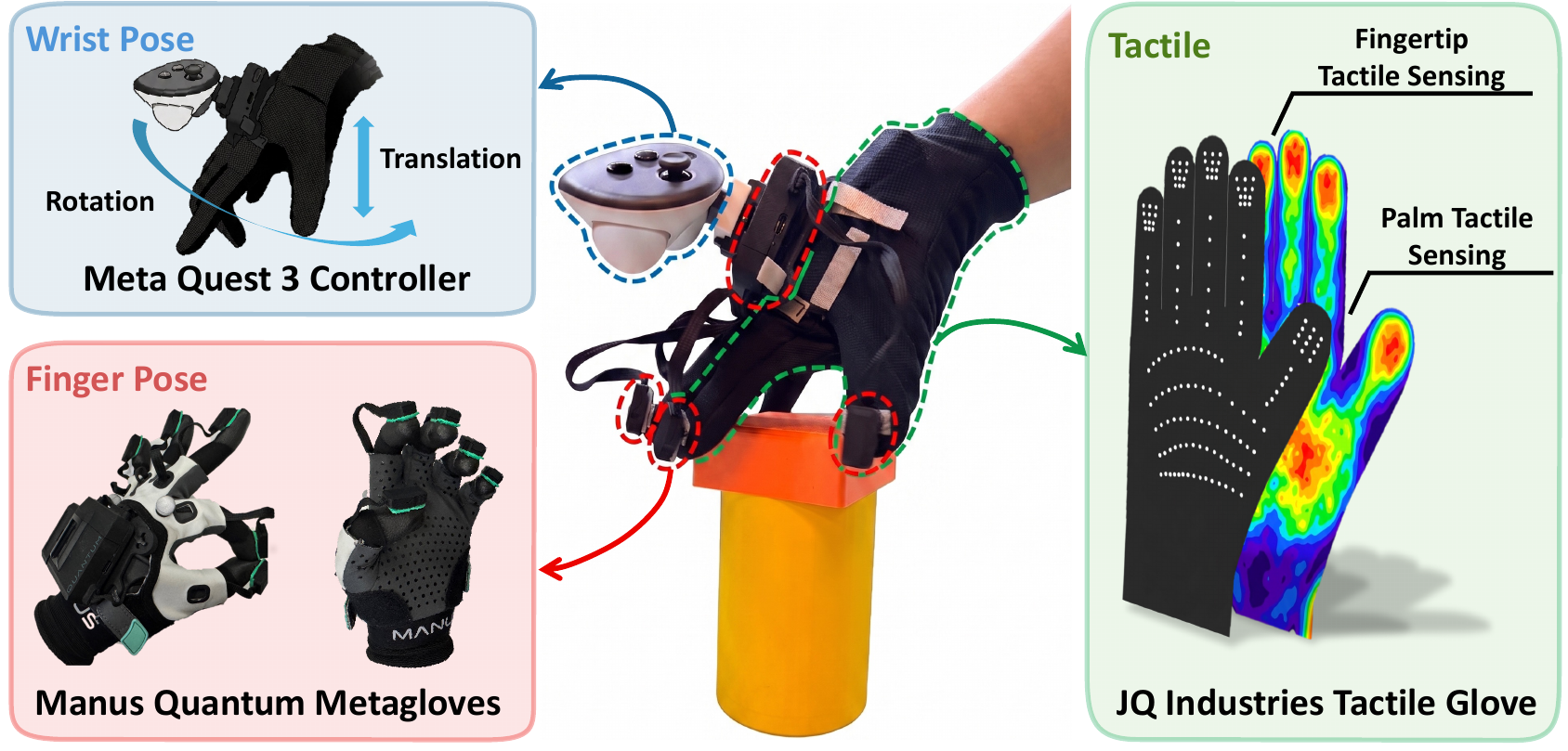}
\caption{\textbf{Multimodal data collection system.} We collect synchronized hand pose and tactile pressure using a Manus Quantum MetaGlove and a JQ-Industries tactile glove.}
\label{fig:data_collect}
\end{figure}

High-quality robot demonstrations of dexterous rotation are difficult to collect due to the intractability of accurately conveying contact perception during teleoperation~\cite{qin2023anyteleop,luo_hajl}. We therefore collect human demonstrations instead by having participants directly rotate the target objects via a tactile glove. For this purpose, we develop the data-collection system shown in Fig.\ref{fig:data_collect}. A Meta Quest 3 controller~\cite{metaquest3} attached to the back of the hand tracks the global hand pose. A Manus Quantum MetaGlove~\cite{manus_vr_glove} captures finger joint poses, while a JQ-Industries tactile glove~\cite{juqiao_industrial} measures contact pressure across the fingertips using a piezoresistive pressure array. Using this device, We concurrently extract fingertip pressure measurements, as well as the hand pose, and RGB observations. In total, we collect over \(100{,}000\) frames of human rotation data. These synchronized recordings allow us to determine active contact states and compute the corresponding contact-conditioned manipulability labels used for pretraining.

\subsection{3.2 Analytic Manipulability from Human}
\label{sec:pretrain}

Using collected human data, we calculate the contact-conditioned manipulability introduced as follows. The overall pipeline for such calculation comprises three components: (1)~a visual--tactile encoder that fuses RGB and tactile histories into a compact representation, (2)~contact-conditioned manipulability, an analytic, tactile-gated manipulability descriptor that provides supervision labels from demonstrations, and (3)~an auxiliary prediction decoder that is discarded after pretraining. We describe each component below.

\subsubsection{Visual--Tactile Encoder.}
Given a human demonstration sequence, the encoder maps these observations to
\begin{equation}
    (z_t^{VT},h_v,h_c)
    =
    F_{\eta}(V_{t-H+1:t},T_{t-H+1:t}),
\end{equation}
where visual and tactile histories are denoted as \(V_{t-H+1:t}\) and \(T_{t-H+1:t}\), respectively, \(H\) is the history length,
\(z_t^{VT}\) is the fused visual--tactile representation, and \(h_v\) and \(h_c\) are modality-specific token features. Using this encoder, the RGB observations are embedded as visual patch tokens, while the tactile readings are projected into contact tokens. A learnable transformer-based~\cite{vaswani2023attentionneed} fusion network \(F_{\eta}\) then integrates the two modalities. 
Thus, \(z_t^{VT}\) compactly describes both the visually observed hand--object interaction and where contact is currently occurring.

\subsubsection{Contact-conditioned Manipulability.}
\label{sec:pretrain:manip}

We introduce contact-conditioned manipulability as a cross-embodiment representation of rotational capability evolution. Conventional configuration based manipulability indices characterize a hand's kinematic capabilities but do not explicitly capture how active contacts constrain object motion. To address this limitation, we introduce contact-conditioned manipulability defined as following.
Let $J_{c,t}^{\mathrm{hum}}$ denote the stacked Jacobian of the active fingertip contacts, where $W_t$ denote the diagonal matrix of tactile-confidence weights, $H_h$ is a diagonal joint-space scaling matrix, and $G_t^+$ the damped pseudoinverse of the grasp matrix. Extending the conventional manipulability formulation of~\cite{chiacchio1991global} with tactile-induced kinematic constraints, we further define the contact capability $C_{c,t}^{\mathrm{hum}}$ and rotational manipulability matrix $M_{\omega,t}^{\mathrm{hum}}$ as
\begin{equation}
\begin{aligned}
C_{c,t}^{\mathrm{hum}}
&=
W_t^{1/2}
J_{c,t}^{\mathrm{hum}}
H_h
(J_{c,t}^{\mathrm{hum}})^\top
W_t^{1/2},\\
M_{\omega,t}^{\mathrm{hum}}
&=
P_\omega
(G_t^+)^\top
C_{c,t}^{\mathrm{hum}}
G_t^+
P_\omega^\top
+
\epsilon_m I_3 .
\end{aligned}
\end{equation}
where \(P_\omega\) selects the rotational component and \(\epsilon_m I_3\) ensures positive
definiteness. We represent \(M_{\omega,t}^{\mathrm{hum}}\) with the
log-Euclidean coordinate
$
m_t^{\mathrm{hum}}
=
\operatorname{vech}
\left(
\log M_{\omega,t}^{\mathrm{hum}}
\right)
\in\mathbb R^6 .
$
Although $m_t$ is computed using each hand's own contact Jacobian, we define the wrist-frame axes consistently across hands, so that its dimensions describe rotational capabilities about the same $x$-, $y$-, and $z$-directions. Contact positions are measured relative to the weighted center of the tactilely detected contacts, avoiding the need to estimate the object center. This shared task-space definition allows human evolution prior to be transferred to robot-induced changes without joint-level correspondence.

\subsection{3.3 Manipulability-Evolution Energy Prior}
\label{sec:energy}

Using the contact-conditioned manipulability descriptors extracted from human demonstrations, we learn a prior over the evolution of manipulability during successful object rotation. The model is capable of evaluating candidate directions of manipulability change, assigning low energy to directions consistent with human demonstrations and high energy to incompatible directions. 

To train this prior model, we use three complementary components: (1) a contrastive energy objective, (2) structured negative sampling, and (3) an auxiliary objective for predicting the magnitude of the change. All these components are detailed as following. 

\subsubsection{Contrastive Manipulability-Evolution Energy.}
We define a conditional energy model
$E_\theta(\hat v\mid c_t)$
that measures the compatibility between a candidate manipulability-change direction \(\hat v\) and the current interaction context
\(
c_t=[z_t^{VT},m_t,d_t]
\),
where \(z_t^{VT}\) encodes the visual--tactile observations, \(m_t\) denotes the current manipulability state, and \(d_t\) specifies the desired rotation direction. A lower energy indicates that the candidate direction is more consistent with the manipulability evolution observed in human demonstrations.

We obtain the positive direction from the demonstrated short-horizon change:
\begin{equation}
\hat v_t^+
=
\frac{
m_{t+\Delta}-m_t
}{
\lVert m_{t+\Delta}-m_t\rVert_2+\epsilon_{\mathrm{num}}
}.
\end{equation}
This produces a unit vector in six-dimensional manipulability space, allowing the energy model to focus on the direction in which manipulability should evolve. We exclude near-stationary frames from the directional objective because their normalized directions are dominated by measurement noise.

\subsubsection{Structured Negatives and Energy Loss.}
To teach the model which directions are incompatible with the current context, we follow the sampling-based contrastive strategy of IBC~\cite{florence2021implicitbehavioralcloning}. We contrast each demonstrated direction with \(K\) structured negatives. These negatives comprise random directions, reversed directions, and directions drawn from mismatched coordinate frames. Random negatives provide broad coverage of the direction space, whereas reversed and mismatched directions provide incorrect instances within interaction context.

Let
\(
\mathcal V_t
=
\{\hat v_t^+\}\cup\{\hat v_{t,k}^-\}_{k=1}^{K}
\)
denote the resulting candidate set. We optimize the contrastive energy loss
\begin{equation}
\mathcal L_E
=
-\log
\frac{
\exp\!\left[-E_\theta(\hat v_t^+\mid c_t)/\tau\right]
}{
\displaystyle
\sum_{\hat v\in\mathcal V_t}
\exp\!\left[-E_\theta(\hat v\mid c_t)/\tau\right]
},
\end{equation}
where \(\tau\) is the contrastive temperature. This objective encourages the demonstrated direction to have lower energy than incompatible alternatives, thereby shaping an energy landscape that captures context-dependent manipulability evolution.

\subsubsection{Magnitude Prediction and Full Pretraining Loss.}
In addition to the aforementioned directional target (normalized, it does not preserve the magnitude of desired motion), we introduce an additional auxiliary head that predicts the log-scaled magnitude:
\begin{equation}
\alpha_t
=
\log\!\left(
1+\lVert m_{t+\Delta}-m_t\rVert_2
\right),
\qquad
\mathcal L_\alpha
=
\lvert\hat\alpha_t-\alpha_t\rvert.
\end{equation}
Thus, the complete pretraining objective is
\begin{equation}
\mathcal L_{\mathrm{pre}}
=
\lambda_{\mathrm{rec}}\mathcal L_{\mathrm{rec}}
+
\lambda_{\mathrm{pred}}\mathcal L_{\mathrm{pred}}
+
\lambda_E\mathcal L_E
+
\lambda_{\alpha}\mathcal L_{\alpha},
\end{equation}
where \(\mathcal L_{\mathrm{rec}}\) and \(\mathcal L_{\mathrm{pred}}\) train the auxiliary decoder and learned representation, respectively, while \(\mathcal L_E\) and \(\mathcal L_\alpha\) supervise the direction and magnitude of manipulability evolution. After pretraining, the visual--tactile encoder \(F_\eta\) and the energy model \(E_\theta\) are frozen and subsequently used during robot reinforcement learning.

\subsection{3.4 Manipulability Energy-Guided Residual Policy}
\label{sec:residual}

Finally, we transfer the manipulability-evolution preference learned from human demonstrations to robot control. Rather than directly executing the direction preferred by the energy model, we use it as a local action-space hint for a residual policy. This design preserves the task-solving capability of the base policy while biasing exploration toward actions that induce human-like, contact-aware manipulability evolution.

\noindent\textbf{Predicting Action-Induced Manipulability Change.}
At time $t$, the robot receives an RGB observation $V_t^r$, tactile signals $T_t^r$, and joint state $q_t$. The frozen visual--tactile encoder produces
\begin{equation}
z_t^r = F_\eta(V_t^r,T_t^r),
\end{equation}
and the base policy proposes a nominal continuous action
\begin{equation}
a_t^0
\sim
\pi_{\mathrm{base},\psi}
\left(
\cdot \mid z_t^r,q_t
\right).
\end{equation}
We represent the robot's current rotational manipulability using the same descriptor as for the human demonstrations:
$
m_t^r
=
\operatorname{vech}
\left(
\log M_{\omega,t}^r
\right)
\in \mathbb{R}^6,
$
where $\operatorname{vech}(\cdot)$ stacks the six unique entries of a symmetric $3\times3$ matrix.

To evaluate how a candidate action changes manipulability, let $u_t(a,q_t)$ denote the local joint displacement induced by action $a$ under the robot controller. We define the resulting normalized manipulability-change direction as
\begin{equation}
\hat{v}_t^r(a)
=
\frac{
m^r\!\left(q_t+u_t(a,q_t)\right)-m_t^r
}{
\left\|
m^r\!\left(q_t+u_t(a,q_t)\right)-m_t^r
\right\|_2+\epsilon
}.
\label{eq:robot_manip_direction}
\end{equation}
This one-step estimate keeps the current contact geometry fixed, providing an efficient local approximation without requiring contact-dynamics rollouts. Contact changes are incorporated at the next control step through the updated tactile observations and robot state. 
% \textcolor{red}{During construction of the grasp matrix, we approximate the object center as the center of the contacting fingertips.}

\noindent\textbf{Energy-Guided Action Bias.}
We define the robot-side energy context as
$
c_t^r=[z_t^r,m_t^r,d_t]
$,
where $d_t$ denotes the desired task rotation axis. To identify a locally preferred action without replacing the base policy, we construct a candidate set around its nominal action:
\begin{equation}
\begin{gathered}
\mathcal{A}_t
=
\{a_t^0\}
\cup
\left\{
\operatorname{clip}_{\mathcal{A}}(a_t^0+\xi_{t,k})
\right\}_{k=1}^{K},\\
\xi_{t,k}
\sim
\mathcal{N}(0,\sigma_a^2 I).
\end{gathered}
\label{eq:local_action_candidates}
\end{equation}
Each candidate is mapped to its induced manipulability-change direction through Eq.~\eqref{eq:robot_manip_direction} and evaluated by the frozen energy model. The lowest-energy candidate then defines a local action-space bias:
\begin{equation}
\begin{gathered}
a_t^\star
=
\arg\min_{a\in\mathcal{A}_t}
E_\theta
\left(
\hat{v}_t^r(a)\mid c_t^r
\right),\\
b_t^E
=
\operatorname{sg}
\left[
\operatorname{clip}_{b_{\max}}
\left(
a_t^\star-a_t^0
\right)
\right].
\end{gathered}
\label{eq:energy_action_bias}
\end{equation}
Here, $\operatorname{sg}(\cdot)$ denotes the stop-gradient operator, and $b_t^E$ indicates which local deviation from the nominal action is predicted to yield a more desirable manipulability evolution. Clipping limits the influence of this guidance and prevents the energy model from overriding the base policy.

\noindent\textbf{Residual Policy and Training.}
Because the one-step energy score captures only a local preference and does not account for long-horizon task outcomes, we do not execute $a_t^\star$ directly. Instead, we condition a learned residual policy on the energy-guided bias:
\begin{equation}
\begin{gathered}
\Delta a_t
\sim
\pi_{\mathrm{res},\phi}
\left(
\cdot
\mid
z_t^r,
q_t,
a_t^0,
m_t^r,
\operatorname{sg}(b_t^E)
\right),\\
a_t
=
\operatorname{clip}_{\mathcal{A}}
\left(
a_t^0+\lambda_R\Delta a_t
\right).
\end{gathered}
\label{eq:energy_residual_policy}
\end{equation}
where $\lambda_R$ controls the magnitude of the residual correction, and $\operatorname{clip}_{\mathcal{A}}(\cdot)$ projects the final command onto the valid action space $\mathcal{A}$. 

During robot learning, the encoder $F_\eta$ and energy model $E_\theta$ remain frozen, while the base and residual policies are optimized solely using the original task reward. Notably, our framework guides policy learning through an embodiment-agnostic preference over manipulability evolution, which differs from conventional approaches that use human demonstrations for hand-specific action imitation, reward shaping, or online optimization. Additional definitions, hyperparameters, and implementation details are provided in the supplementary material.

%% file: Tex/4_exp.tex
We evaluate whether contact-conditioned manipulability evolution provides effective and reusable guidance for dexterous rotation. We first measure task success and generalization to held-out objects across three LEAP Hand tasks, as well as cap unscrewing with the Shadow, Allegro, and XHand, using the same frozen human-derived prior. We then conduct controlled ablations to isolate the benefit of state-aligned evolution guidance. Finally, we assess execution quality and demonstrate closed-loop deployment in the real world.

\subsection{4.1 Experimental Setup}

We evaluate our approach on three contact-rich rotation tasks: (1) \textit{Unscrew Cap}, (2) \textit{Rotate Object}, and (3) \textit{Turn Faucet}. A trial is considered successful if the cap, object, or faucet handle rotates by at least $2\pi$ about the target axis.

We pretrain the human-derived preference prior using demonstrations of \textit{Unscrew Cap} and \textit{Rotate Object}. To assess cross-task generalization, we exclude \textit{Turn Faucet} from prior pretraining. We further enforce two criteria to evaluate object-level generalization rigorously: (1) objects appearing in the human demonstrations are excluded from robot-policy training and evaluation; and (2) the objects for each robot task are divided into \textit{Seen} objects used for policy training and \textit{Unseen} objects reserved exclusively for evaluation.

We conduct three categories of evaluation. (1) First, to determine whether the proposed prior improves task learning, we train a separate LEAP-hand policy for each task and evaluate its performance. (2) Second, to assess cross-task transfer, we train and evaluate a policy on \textit{Turn Faucet} using the same frozen prior (this task is excluded from prior pretraining). (3) Third, to examine cross-hand transfer, we train Shadow, Allegro, and XHand policies for \textit{Unscrew Cap} using the same task definition, reward function, and frozen preference prior. These evaluations collectively assess whether a single human-derived prior can facilitate policy learning across tasks, objects, and hand morphologies.

We evaluate three aspects of performance: (1) \textit{task success}, (2) \textit{generalization to unseen objects}, and (3) \textit{execution quality}. We report the success rate (SR, \%) separately for \textit{Seen} and \textit{Unseen} objects. \textit{Avg.\ SR} denotes the unweighted mean across all conditions. All results are reported as the mean $\pm$ standard deviation over three independent evaluation seeds. For each seed, we evaluate $1{,}000$ episodes per object with randomized initial hand poses.

\begin{table}[t]
\centering
\small
\setlength{\tabcolsep}{3.5pt}
\begin{tabular}{lccccc}
\toprule
\multirow{2}{*}{\textbf{Method}}
& \multicolumn{2}{c}{\textbf{Unscrew}}
& \multicolumn{2}{c}{\textbf{Faucet}}
& \multirow{2}{*}{\textbf{Avg. SR}} \\
\cmidrule(lr){2-3}
\cmidrule(lr){4-5}
& \textbf{Seen}
& \textbf{Unseen}
& \textbf{Seen}
& \textbf{Unseen}
& \\
\midrule

Zero Guidance
& $18.6 $
& $9.4 $
& $48.6$
& $20.2 $
& 24.2 \\

Context-Shuffled
& $14.0 $
& $5.3 $
& $22.5 $
& $7.3 $
& 12.3 \\

Greedy-M
& $54.5 $
& $36.3 $
& $67.8 $
& $44.7 $
& 50.8 \\

\rowcolor[gray]{0.93}
\textbf{DexMani}
& $\mathbf{65.1 }$
& $\mathbf{39.3 }$
& $\mathbf{80.2 }$
& $\mathbf{70.3 }$
& $\mathbf{63.7}$ \\

\bottomrule
\end{tabular}

\caption{\textbf{Ablation Studies.}
All methods use the same training setup, differing only in the guidance signal.}
\label{tab:ablation}
\end{table}

\begin{table*}[ht]
\centering
\small
\setlength{\tabcolsep}{3.5pt}
\begin{tabular}{l *{3}{ccc}}
\toprule
\multirow{2}{*}{\textbf{Method}} &
\multicolumn{3}{c}{\textbf{Unscrew Cap}} &
\multicolumn{3}{c}{\textbf{Rotate Object}} &
\multicolumn{3}{c}{\textbf{Turn Faucet}} \\
\cmidrule(lr){2-4}
\cmidrule(lr){5-7}
\cmidrule(lr){8-10}
& TCI $\uparrow$ & LDLJ $\uparrow$ & SPARC $\uparrow$
& TCI $\uparrow$ & LDLJ $\uparrow$ & SPARC $\uparrow$
& TCI $\uparrow$ & LDLJ $\uparrow$ & SPARC $\uparrow$ \\
\midrule
PPO
& 0.29 $\pm {\scriptstyle 0.11}$ & -25.6 $\pm {\scriptstyle 1.4}$ & -6.0 $\pm {\scriptstyle 0.5}$
& $\mathbf{0.43} \pm {\scriptstyle 0.11}$ & -14.5 $\pm {\scriptstyle 0.1}$ & -2.6 $\pm {\scriptstyle 0.2}$
& 0.22 $\pm {\scriptstyle 0.05}$ & -23.2 $\pm {\scriptstyle 2.3}$ & -4.8 $\pm {\scriptstyle 0.8}$ \\

VT Pretrain
& 0.41 $\pm {\scriptstyle 0.13}$ & -24.5 $\pm {\scriptstyle 1.1}$ & -5.5 $\pm {\scriptstyle 0.6}$
& 0.41 $\pm {\scriptstyle 0.23}$ & -14.4 $\pm {\scriptstyle 0.1}$ & -2.8 $\pm {\scriptstyle 0.2}$
& 0.23 $\pm {\scriptstyle 0.06}$ & -23.2 $\pm {\scriptstyle 2.0}$ & -4.5 $\pm {\scriptstyle 0.9}$ \\

VTM
& 0.44 $\pm {\scriptstyle 0.17}$ & -24.1 $\pm {\scriptstyle 0.8}$ & -5.3 $\pm {\scriptstyle 0.7}$
& 0.26 $\pm {\scriptstyle 0.18}$ & -15.5 $\pm {\scriptstyle 0.8}$ & -3.3 $\pm {\scriptstyle 0.3}$
& 0.27 $\pm {\scriptstyle 0.15}$ & -23.4 $\pm {\scriptstyle 1.8}$ & -4.4 $\pm {\scriptstyle 0.4}$ \\

VTA
& 0.45 $\pm {\scriptstyle 0.12}$ & -23.7 $\pm {\scriptstyle 1.6}$ & -5.0 $\pm {\scriptstyle 0.1}$
& 0.29 $\pm {\scriptstyle 0.09}$ & -15.6 $\pm {\scriptstyle 0.9}$ & -3.1 $\pm {\scriptstyle 0.2}$
& 0.23 $\pm {\scriptstyle 0.04}$ & -23.9 $\pm {\scriptstyle 2.4}$ & -4.1 $\pm {\scriptstyle 1.2}$ \\

VTA-E
& 0.43 $\pm {\scriptstyle 0.05}$ & -23.6 $\pm {\scriptstyle 0.4}$ & -5.0 $\pm {\scriptstyle 0.3}$
& 0.23 $\pm {\scriptstyle 0.10}$ & -13.9 $\pm {\scriptstyle 0.2}$ & -2.8 $\pm {\scriptstyle 0.5}$
& 0.20 $\pm {\scriptstyle 0.02}$ & -24.8 $\pm {\scriptstyle 0.4}$ & -4.5 $\pm {\scriptstyle 0.8}$ \\

\rowcolor[gray]{0.93}
\textbf{DexMani}
& $\mathbf{0.50} \pm {\scriptstyle 0.13}$ & $\mathbf{-23.1} \pm {\scriptstyle 0.5}$ & $\mathbf{-4.5} \pm {\scriptstyle 0.4}$
& 0.36 $\pm {\scriptstyle 0.05}$ & $\mathbf{-11.9} \pm {\scriptstyle 1.8}$ & $\mathbf{-2.3} \pm {\scriptstyle 0.7}$
& $\mathbf{0.28} \pm {\scriptstyle 0.07}$ & $\mathbf{-22.4} \pm {\scriptstyle 0.5}$ & $\mathbf{-3.9} \pm {\scriptstyle 0.1}$ \\
\bottomrule
\end{tabular}
\caption{Quantitative comparison of execution quality and efficiency across three dexterous rotation tasks on the LEAP hand.}
\label{tab:dexterity}
\end{table*}

\subsection{4.2 Performance of the Proposed Approach}

\noindent\textbf{Generalization across tasks and objects.}
We compare DexMani against five baselines: (1) \textit{PPO}, which learns solely from the task reward; (2) \textit{VT Pretraining}, which learns a visual--tactile policy representation from human demonstrations via pretraining, without manipulability supervision or online guidance; (3) \textit{VTM}, which additionally predicts contact-conditioned manipulability during pretraining, but does not utilize manipulability during policy learning; (4) \textit{VTA}, which additionally guides policy learning using retargeted human actions; and (5) \textit{VTA-E}, which additionally learns an energy model over retargeted actions and provides guidance through the same residual interface as DexMani. The details, and additional differences between these baselines are reported in the supplementary material. 

The results are reported in Table~\ref{tab:task_transfer} and reveal three main findings. First, DexMani achieves the highest SR across all six task--split conditions, attaining an Avg.\ SR of 57.5\% and outperforming VTM, the strongest baseline, by 5.6 percentage points. Second, the improvements are particularly pronounced on \textit{Unseen} objects, indicating that the learned guidance generalizes beyond the object instances and categories encountered during policy training. Third, DexMani performs strongly on \textit{Turn Faucet}, despite this task being excluded from human-prior pretraining. This result demonstrates that the same frozen prior can transfer to a previously unseen rotation task.

The baseline comparisons provide two additional insights. First, the limited performance of VT Pretraining and VTM indicates that representation pretraining on human demonstrations alone, even with manipulability supervision, is insufficient to fully exploit the transferred knowledge without online policy guidance. Second, the performance of VTA and VTA-E suggests that retargeted human actions transfer less effectively across embodiments than the proposed manipulability-based representation. In contrast, DexMani transfers a preference over how rotational manipulability should evolve, allowing each robot to realize this preference according to its own morphology, and kinematic constraints. Consequently, a single prior can guide multiple rotation tasks while generalizing to previously unseen objects.

\noindent\textbf{Generalization across hands.}
To assess whether the learned prior transfers across embodiments, we evaluate DexMani on three dexterous hands: (1) Shadow, (2) Allegro, and (3) XHand. For all three embodiments, we use the same \textit{Unscrew Cap} task, reward function, and preference prior model, while training a separate policy in each hand's native action space.
As shown in Table~\ref{tab:hand_transfer}, DexMani achieves the highest SR on both \textit{Seen} and \textit{Unseen} objects for all three hands. These consistent improvements demonstrate that manipulability-based guidance transfers across distinct hand kinematics.

\begin{figure}[t]
\centering
\includegraphics[width=0.47\textwidth]{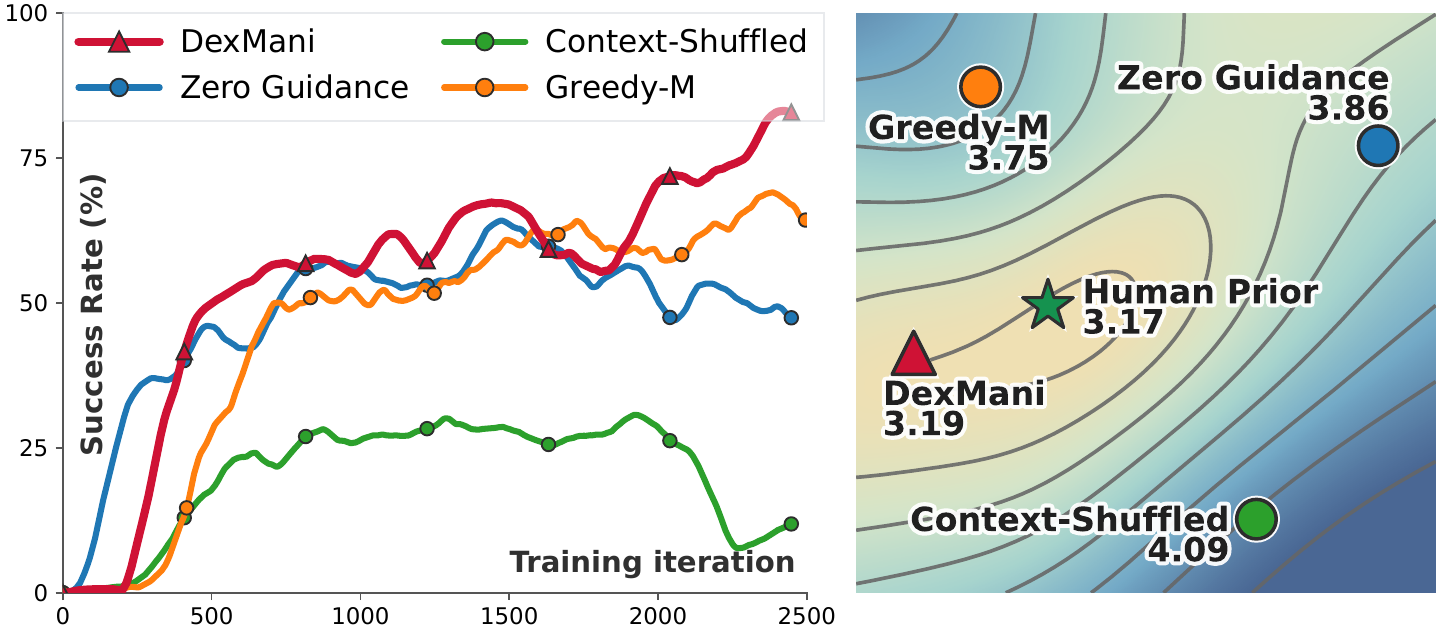}
\caption{\textbf{Learning and energy analysis.} Left: Turn Faucet learning curves. Right: Projected action-energy landscape and selected actions under different guidance strategies.}
\label{fig:mechanism}
\end{figure}

\subsection{4.3 Ablation Study}

To understand why the manipulability preference prior improves policy performance, we investigate three questions: (1) Does manipulability-based guidance improve action selection? (2) If so, must this guidance be conditioned on the correct interaction context? (3) Is guiding a residual policy using the short-horizon evolution of human manipulability preferences more effective than greedily maximizing the robot's instantaneous rotational manipulability? We address these questions using three ablations: \textit{Zero Guidance}, which removes guidance from the residual policy's input; \textit{Context-Shuffled}, which replaces the correct guidance with mismatched interaction context; and \textit{Greedy-$M$}, replaces the learned evolution guidance with a bias toward the candidate that maximizes instantaneous rotational manipulability, while retaining the same residual-policy interface.

As shown in Table~\ref{tab:ablation}, DexMani achieves the highest success rate across all conditions. Its improvements over \textit{Zero Guidance} and \textit{Context-Shuffled} confirm the benefits of manipulability-based guidance matched to the interaction context. The advantage over \textit{Greedy-$M$} shows that the gain does not come from simply increasing instantaneous rotational capability. Instead, guidance learned from successful manipulability evolution provides a more effective signal for sustained rotation. Consistently, Figure~\ref{fig:mechanism} shows that DexMani learns faster and performs best on \textit{Turn Faucet}. The action--energy landscape further reveals that DexMani selects an action in a low-energy region, whereas \textit{Greedy-$M$} is only partially aligned with the human-derived prior. These results demonstrate the importance of interaction-conditioned, temporally informed guidance.

\subsection{4.4 Motion Quality and Deployment}

Beyond task success, we further assess execution quality in terms of rotational capability and motion smoothness. We quantify rotational capability using the Task Compatibility Index (TCI)~\cite{chiu1988task}, which measures the ability of the hand--object system to generate motion about the target rotation axis; a higher TCI indicates greater task-aligned rotational capability. We evaluate motion smoothness in both the time and frequency domains using log dimensionless jerk (LDLJ)~\cite{Hogan2009SensitivityOS} and spectral arc length (SPARC)~\cite{balasubramanian2015analysis}, respectively. Higher LDLJ and SPARC values indicate smoother motion.
As shown in Table~\ref{tab:dexterity}, DexMani achieves the best LDLJ and SPARC scores across all three tasks, indicating smoother and more coordinated object motions. It also obtains the highest TCI on \textit{Unscrew Cap} and \textit{Turn Faucet}, demonstrating a stronger ability to generate rotational motion aligned with the target axis.

\begin{figure}[t]
\centering
\includegraphics[width=0.47\textwidth]{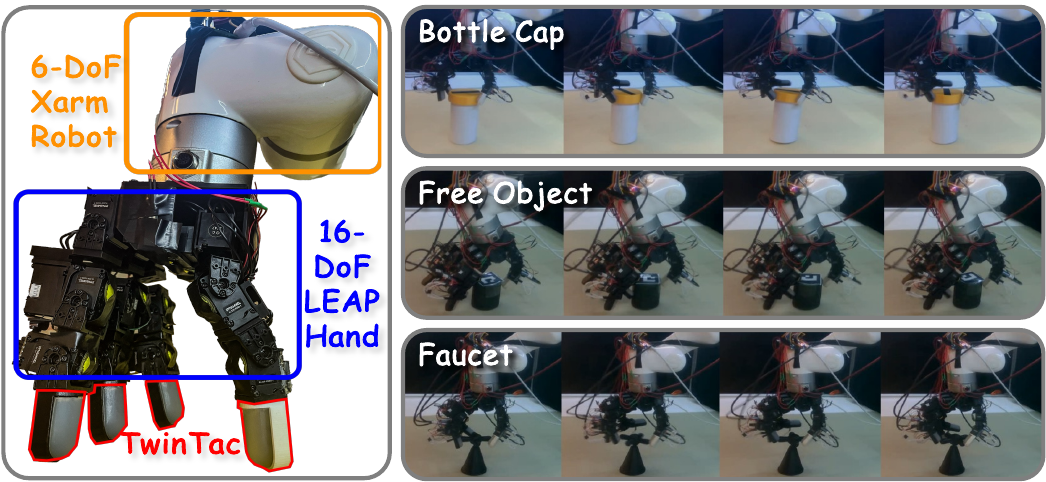}
\caption{\textbf{DexMani real-world deployment.} The sequence shows closed-loop execution at \(20\,\mathrm{Hz}\) in the real world.}
\label{fig:realworld}
\end{figure}

\begin{table}[t]
    \centering
    \small
    \begin{tabular}{lccc}
        \toprule
        \textbf{Method} & \textbf{Unscrew Cap} & \textbf{Rotate Object} & \textbf{Turn Faucet} \\
        \midrule
        DexMani & 6/10 & 3/10 & 1/10 \\
        \bottomrule
    \end{tabular}
    \caption{Real-world DexMani deployment results.}
    \label{tab:real_world_success}
\end{table}

Next, we deploy DexMani on a physical system consisting of a 16-DoF LEAP Hand~\cite{shaw2023leaphand} mounted on a 6-DoF xArm, as shown in Figure~\ref{fig:realworld}. The fingertips are equipped with TwinTac tactile sensors~\cite{huang2025twintac}, whose measurements are binarized to indicate contact. At each control step, the policy integrates these tactile signals with visual and proprioceptive observations to predict joint-position targets. To facilitate sim-to-real transfer, we apply domain randomization~\cite{tobin2017domainrandomizationtransferringdeep} during simulation training, improving robustness to variations in physical dynamics and sensory observations. As reported in Table~\ref{tab:real_world_success}, DexMani successfully performs closed-loop rotation tasks on the physical system, demonstrating its ability to transfer from simulation to the real world despite discrepancies in sensing and dynamics.

%% file: Tex/5_conlcusion.tex
We introduced DexMani, a framework that transfers human rotational manipulation experience through contact-conditioned manipulability evolution. By representing human preferences as manipulability evolution rather than hand-specific joint trajectories, DexMani improves task success, motion smoothness, and rotational capability while generalizing across tasks and objects. We also demonstrated its successful deployment on a physical LEAP Hand. These results establish manipulability evolution as a transferable representation of human experience across tasks, objects, and hand morphologies.
One main limitation of this work is that, although the prior is shared, each task–hand configuration still requires a separately trained policy, precluding direct transfer of a unified policy across embodiments. Future work will focus on improving Sim2Real performance and developing unified policies that support direct cross-embodiment transfer.

%% file: sup_arxiv.tex
\lstset{%
	basicstyle={\footnotesize\ttfamily},% footnotesize acceptable for monospace
	numbers=left,numberstyle=\footnotesize,xleftmargin=2em,% show line numbers, remove this entire line if you don't want the numbers.
	aboveskip=0pt,belowskip=0pt,%
	showstringspaces=false,tabsize=2,breaklines=true}
\floatstyle{ruled}
\newfloat{listing}{tb}{lst}{}
\floatname{listing}{Listing}

\pdfinfo{
/TemplateVersion (2027.1)
}

\setcounter{secnumdepth}{0} 

\title{Supplementary Material for DexMani}
\author{
    Written by AAAI Press Staff\textsuperscript{\rm 1}\thanks{With help from the AAAI Publications Committee.}\\
    AAAI Style Contributions by Peter Patel Schneider,
    Sunil Issar,\\
    J. Scott Penberthy,
    George Ferguson,
    Hans Guesgen,
    Francisco Cruz\equalcontrib\corresponding,
    Marc Pujol-Gonzalez\equalcontrib\corresponding
}
\affiliations{
    \textsuperscript{\rm 1}Association for the Advancement of Artificial Intelligence\\

    1101 Pennsylvania Ave, NW Suite 300\\
    Washington, DC 20004 USA\\
    proceedings-questions@aaai.org

}

\iffalse
\title{My Publication Title --- Single Author}
\author {
    Author Name
}
\affiliations{
    Affiliation\\
    Affiliation Line 2\\
    name@example.com
}
\fi

\iffalse
\title{My Publication Title --- Multiple Authors}
\author {
    First Author Name\textsuperscript{\rm 1,\rm 2}\equalcontrib,
    Second Author Name\textsuperscript{\rm 2}\equalcontrib,
    Third Author Name\textsuperscript{\rm 1}\corresponding
}
\affiliations {
    \textsuperscript{\rm 1}Affiliation 1\\
    \textsuperscript{\rm 2}Affiliation 2\\
    firstAuthor@affiliation1.com, secondAuthor@affilation2.com, thirdAuthor@affiliation1.com
}
\fi

\overfullrule=5pt

\maketitle

\section{1 Human Dataset and Hardware Details}
\label{app:human-dataset}

\label{app:method-details}

Human demonstrations provide supervision for both the visual--tactile encoder and the prior model of contact-conditioned rotational manipulability evolution. To explain how these data are collected, we briefly describe the data collection system and the processing procedure of the dataset.

\begin{figure}[ht]
    \centering
    \includegraphics[width=0.85\linewidth]{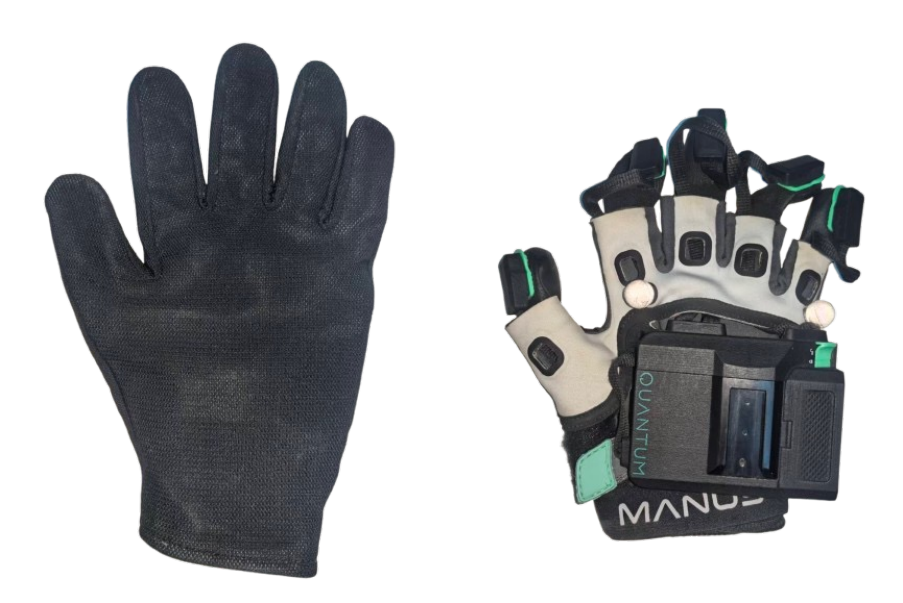}
    \caption{\textbf{Hardware for human data collection.}
    Left: the Juqiao tactile glove. Right: the Manus Quantum MetaGlove.}
    \label{fig:Hardwares}
\end{figure}

\paragraph{Data collection system.}
We collect synchronized multimodal recordings of human object rotation using a wearable acquisition system. The system captures RGB video, right-hand kinematics, and dense tactile measurements at 30~Hz. Hand kinematics are recorded using a Manus Quantum MetaGlove (Fig.~\ref{fig:Hardwares}). The Manus hand model provides the three-dimensional positions of 21 hand joints. Contact signals are captured using a 256-channel Juqiao piezoresistive tactile glove. At each frame, the tactile glove outputs a pressure vector $u_t \in \{0,\ldots,255\}$, where each element corresponds to a tactile taxel and larger values indicate stronger local pressure responses. Two RGB cameras record the hand--object interaction at a resolution of $640 \times 480$. All sensor streams are temporally aligned using timestamps. 

\begin{figure}[t]
    \centering
    \includegraphics[width=0.45\textwidth]{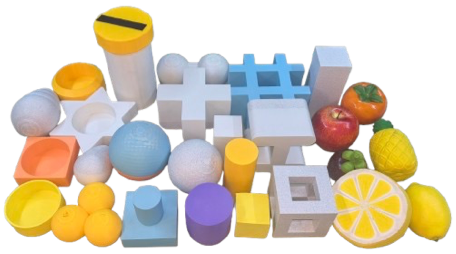}
    \caption{Objects used for human demonstration collection.}
    \label{fig:dataset_objects}
\end{figure}

As shown in Fig.~\ref{fig:dataset_objects}, the human-demonstration dataset contains 27 objects, including common geometric primitives and everyday household items. This collection encompasses diverse object geometries and contact patterns for learning the evolution of rotational manipulability.

\paragraph{Visual processing.}
We use the recorded human-hand trajectories to animate a MANO hand model~\cite{MANO:SIGGRAPHASIA:2017}. The resulting hand motions are rendered in simulation to provide additional RGB observations. During pretraining, both the captured RGB images and the MANO-rendered images are used as visual inputs.

\paragraph{Tactile processing.}
We partition the raw tactile taxels into five semantic fingertip regions corresponding to the thumb, index, middle, ring, and little fingers. For each finger $i$, we aggregate the responses of the taxels assigned to its fingertip region. The corresponding taxel set, $\mathcal{S}_i$, is determined by the physical layout of the Juqiao tactile glove. The finger-level tactile response is defined as
\begin{equation}
f_{i,t}
=
\sum_{\ell \in \mathcal{S}_i}
\left|u_{\ell,t}\right|.
\end{equation}

\begin{figure*}[t]
\centering
\includegraphics[width=0.97\textwidth]{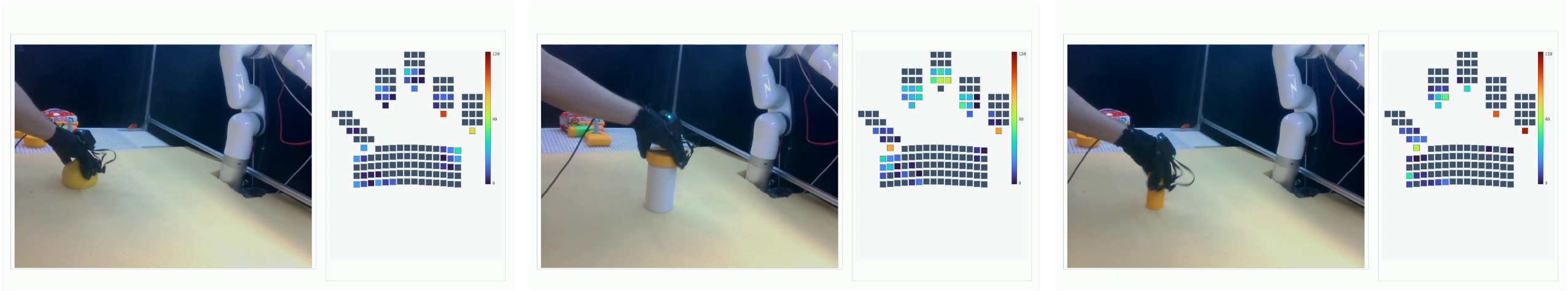}
\caption{\textbf{Human Dataset Visualization.} Each pair shows the RGB observation and the corresponding tactile-pressure map. }
\end{figure*}

We compute a contact threshold separately for each recording $r$:
\begin{equation}
\tau_r
=
0.15\,
\operatorname{Percentile}_{95}
\left(
\left\{f_{i,t}
\mid i \in \{1,\ldots,5\},\, t \in r\right\}
\right).
\end{equation}
The binary contact state is then given by
\begin{equation}
c_{i,t}
=
\mathbb{I}
\left[
f_{i,t} > \tau_r
\right].
\end{equation}
Only fingertips for which $c_{i,t}=1$ are included when constructing the contact Jacobian and grasp matrix~\cite{graspmatrix}. For visual--tactile pretraining, the tactile input to the encoder is
\begin{equation}
T_t
=
\left[
(c_{1,t}),\ldots,
(c_{5,t})
\right].
\end{equation}
Thus, the encoder receives five finger-level tokens rather than the raw tactile vector.

During RL training, we do not match individual taxels across different hands. Instead, we group the tactile sensors by finger and convert each finger's response into a binary contact state. We use the same contact threshold for all robot hands, setting the simulated force threshold to $0.01\,\mathrm{N}$. For four-fingered hands, the little-finger entry is set to zero. Consequently, the method requires only finger-level contact information and does not rely on identical sensor layouts, taxel-level correspondence, or contact-normal estimation.

\section{2 Additional Method Details}

This section provides the mathematical details omitted from the main text for clarity. We first describe how human manipulability labels are analytically computed from hand kinematics and tactile contacts. We then detail the manipulability-evolution energy objective and the robot-side realization of the residual action guidance.

\subsection{2.1 Contact-conditioned Manipulability}
\label{app:human-manip}

We compute contact-conditioned manipulability from the human hand configuration and the tactile contact state. The computation contains three steps. We first construct the human hand Jacobian from the Manus hand model. Then we modulate the Jacobians with the active contact information. Finally, we map the resulting motion capability to the object rotational space.

\paragraph{Human hand Jacobian.}
The Manus glove provides hand poses at each frame. Each finger is represented as a serial kinematic chain. For an active fingertip contact \(i\), its positional Jacobian is obtained by differentiating the forward kinematics of the corresponding finger:
\begin{equation}
J_{i,t}
=
\frac{\partial p_{i,t}}{\partial q_t},
\end{equation}
where \(q_t\) denotes the human hand configuration and \(p_{i,t}\) denotes the fingertip position. We stack the Jacobians of all active contacts into
\begin{equation}
J_{c,t}^{\mathrm{hum}}
=
\begin{bmatrix}
J_{1,t}\\
\vdots\\
J_{N_t,t}
\end{bmatrix},
\qquad
\dot p_t = J_t\dot q_t,
\label{eq:human_contact_jacobian}
\end{equation}
where \(N_t\) is the number of active contacts.

We compute the human contact-conditioned rotational manipulability as
\begin{equation}
\begin{aligned}
C_{c,t}^{\mathrm{hum}}
&=
W_t^{1/2}
J_{c,t}^{\mathrm{hum}}
H_h
(J_{c,t}^{\mathrm{hum}})^\top
W_t^{1/2},\\
M_{\omega,t}^{\mathrm{hum}}
&=
P_\omega
(G_t^+)^\top
C_{c,t}^{\mathrm{hum}}
G_t^+
P_\omega^\top
+
\epsilon_m I_3 .
\end{aligned}
\label{eq:human_contact_manip}
\end{equation}
The terms in Eq.~\eqref{eq:human_contact_manip} are defined below.

\paragraph{Contact activation matrix $W_t$.}
In the general formulation, $W_t$ is a block-diagonal matrix that removes inactive fingertip contacts. In our implementation, inactive contacts are removed before constructing $J_{c,t}^{\mathrm{hum}}$ and $G_t$. Therefore,
\begin{equation}
W_t=I_{3N_t},
\end{equation}
and no additional contact masking is applied.

\paragraph{Joint-space scaling matrix \(H_h\).}
The matrix
\(
H_h=\operatorname{diag}(h_1,\ldots,h_{n_h})
\)
controls the contribution of each human joint direction. A larger \(h_j\) gives the \(j\)-th joint direction a larger contribution to \(C_{c,t}^{\mathrm{hum}}\). In our implementation, we set \(H_h=I_{n_h}\) and weight all joint directions equally.

\paragraph{Grasp matrix \(G_t\).}
Let $p_{i,t}$ be the position of active contact $i\in\mathcal{A}_t$ in the wrist frame. We use the center of the active contacts as the local reference point:
\begin{equation}
\bar p_t
=
\frac{1}{N_t}
\sum_{i\in\mathcal{A}_t}p_{i,t}.
\end{equation}
The relative contact position is
\begin{equation}
r_{i,t}=p_{i,t}-\bar p_t.
\end{equation}
The standard grasp matrix is
\begin{equation}
G_t
=
\begin{bmatrix}
I_3 & \cdots & I_3\\
[r_{1,t}]_\times & \cdots & [r_{N_t,t}]_\times
\end{bmatrix},
\end{equation}
where \([\cdot]_\times\) denotes the skew-symmetric matrix. We use the damped
pseudoinverse
\begin{equation}
G_t^+
=
G_t^\top
\left(
G_tG_t^\top+\lambda_G I_6
\right)^{-1}.
\end{equation}
The term \((G_t^+)^\top C_{c,t}^{\mathrm{hum}}G_t^+\) maps the
contact-space motion capability into the object-motion space.
The grasp mapping first produces a \(6\times6\) object-space manipulability
matrix. Its first three dimensions describe translation, and its last three
dimensions describe rotation. We use
\begin{equation}
P_\omega
=
\begin{bmatrix}
0_{3\times3} & I_3
\end{bmatrix}
\end{equation}
to select the rotational component. The resulting
\(M_{\omega,t}^{\mathrm{hum}}\in\mathbb R^{3\times3}\) describes the available
object rotation about the three wrist-frame axes.

Finally, we convert the rotational manipulability matrix into a
six-dimensional descriptor:
\begin{equation}
m_t^{\mathrm{hum}}
=
\operatorname{vech}
\left(
\log M_{\omega,t}^{\mathrm{hum}}
\right)
\in\mathbb R^6.
\end{equation}
DexMani learns the normalized short-horizon changes of
\(m_t^{\mathrm{hum}}\). These changes describe how rotational capability
evolves with the hand configuration and active contacts.

% \textcolor{red}{Q6:Most Critical Comment: The "Embodiment-Agnostic" Representation Is Not Justified.}
\paragraph{Cross-embodiment convention.} Here, we explain why the contact-conditioned manipulability enables cross-embodiment applications. Specifically, the descriptor \(m_t=\operatorname{vech}(\log M_{\omega,t})\) is not invariant to arbitrary coordinate changes. We instead use a shared wrist-frame convention for human demonstrations and robot rollouts. Fingertip positions and Jacobians are expressed in wrist frame before manipulability is computed. In all experiments, the target rotation direction is represented by the hand-frame \(z\)-axis, i.e., \(d=[0,0,1]\). Tactile readings are binarized and used only to select active fingertip contacts. 

After this alignment, the components of \(m_t\) refer to the same wrist-frame rotation axes for both the human and the robot. The human-trained energy model can therefore score the manipulability changes induced by robot actions in the same coordinate system. No joint-level correspondence or action retargeting is required. Each robot produces a preferred change through its own joint motion and contact transitions.

\subsection{2.2 Robot-Side Residual Action}
\label{app:robot-residual}

\begin{algorithm}[ht]
\caption{Robot-side update at one control step.}
\label{alg:supp_update}
\begin{algorithmic}[1]
\REQUIRE Frozen encoder \(F_{\eta}\) and energy model \(E_{\theta}\)
\REQUIRE Robot observation \((V_t^r,T_t^r,q_t)\) and direction \(d_t\)

\STATE Compute the visual--tactile feature
\(z_t^r=F_{\eta}(V_t^r,T_t^r)\).
\STATE Compute the current descriptor
\(m_t^r=\operatorname{vech}(\log M_{\omega,t}^r)\).
\STATE Sample the nominal action
\(a_t^0\sim\pi_{\mathrm{base},\psi}(\cdot\mid z_t^r,q_t)\).
\STATE Construct the local candidate set
\[
\mathcal A_t
=
\{a_t^0\}
\cup
\left\{
\operatorname{clip}_{\mathcal A}(a_t^0+\xi_k)
\right\}_{k=1}^{K}.
\]
\STATE Estimate \(\hat v_t^r(a)\) for each candidate using the current
contact geometry.
\STATE Discard candidates with near-zero predicted manipulability change.
\STATE Set the energy context
\(c_t^r=[z_t^r,m_t^r,d_t]\).
\STATE Score each remaining candidate by
\(S_t(a)=E_{\theta}(\hat v_t^r(a)\mid c_t^r)\).
\STATE Select
\(a_t^\star=\arg\min_{a\in\mathcal A_t}S_t(a)\).
\STATE Compute the stop-gradient bias
\[
b_t^E
=
\operatorname{sg}
\left[
\operatorname{clip}_{b_{\max}}
(a_t^\star-a_t^0)
\right].
\]
\STATE Sample the residual action
\[
\Delta a_t
\sim
\pi_{\mathrm{res},\phi}
(\cdot\mid z_t^r,q_t,a_t^0,m_t^r,b_t^E).
\]
\STATE Execute
\[
a_t
=
\operatorname{clip}_{\mathcal A}
(a_t^0+\lambda_R\Delta a_t).
\]
\STATE Update \(\pi_{\mathrm{base},\psi}\) and
\(\pi_{\mathrm{res},\phi}\) with PPO.
\end{algorithmic}
\end{algorithm}

This section provides the implementation details used in our methods. At each control step, the base policy produces a normalized action
\(a_t^0\). We sample eight nearby actions around
\(a_t^0\). Each candidate action is rescaled to a target joint
configuration in Isaac Gym.

For a candidate \(a\), we compute the joint displacement
\begin{equation}
   u_t(a,q_t)
=
q_t^{\mathrm{tar}}(a)-q_t, 
\end{equation}

where \(q_t\) is the current joint configuration and
\(q_t^{\mathrm{tar}}(a)\) is the target specified by \(a\).
We use the robot fingertip Jacobian to calculate the resulting fingertip motion:
\begin{equation}
\begin{gathered}
J_{\mathrm{tip},t}
=
\frac{\partial p_{\mathrm{tip}}(q_t)}
{\partial q_t},\\
\Delta p_t(a)
=
J_{\mathrm{tip},t}\,u_t(a,q_t).
\end{gathered}
\end{equation}
The candidate contact positions are obtained by adding $\Delta p_t(a)$ to the current positions of the active fingertips, the contact Jacobian is recomputed at the candidate joint
configuration \(q_t^{\mathrm{tar}}(a)\), these contact positions and Jacobian are then used to construct the grasp matrix and the corresponding rotational manipulability
$M_{\omega,t}^{r}(a)$. Finally, we convert
$M_{\omega,t}^{r}(a)$ into its log-Euclidean descriptor and
normalize its difference from the current descriptor to obtain
$\hat{v}_{t}^{r}(a)$, following Eq.~(9) in the main paper. The frozen energy-based model (EBM)~\cite{lecun2006tutorial} scores the manipulability change induced by each of the nine candidates, including the nominal action. A lower energy indicates better agreement with the human-derived manipulability-evolution prior.

\subsection{2.3 Computational Cost.}
We profile one training iteration on an NVIDIA RTX 4090. PPO takes \(5.77\) s per iteration, while DexMani takes \(11.50\) s. The detailed runtime breakdown is shown in Table~\ref{tab:runtime_profile}. The local manipulability computation and EBM scoring together take \(0.434\) s. 

\begin{table}[ht]
    \centering
    \small
    \begin{tabular}{lc}
        \toprule
        \textbf{Method} & \textbf{Time (s)} \\
        \midrule
        PPO & 5.773 \\
        VT Pretraining & 9.256 \\
        VTM & 10.447 \\
        VTA & 9.436 \\
        VTA-E & 10.823 \\
        DexMani & 11.497 \\
        \midrule
        \multicolumn{2}{l}{\textbf{DexMani components}} \\
        \cmidrule(lr){1-2}
        Local manipulability computation & 0.410 \\
        EBM context and candidate scoring & 0.024 \\
        Total local guidance computation & 0.434 \\
        \bottomrule
    \end{tabular}
    \caption{Runtime profiling.}
    \label{tab:runtime_profile}
\end{table}

\section{3 Training Details.}

\begin{figure}[ht]
\centering
\includegraphics[width=0.47\textwidth]{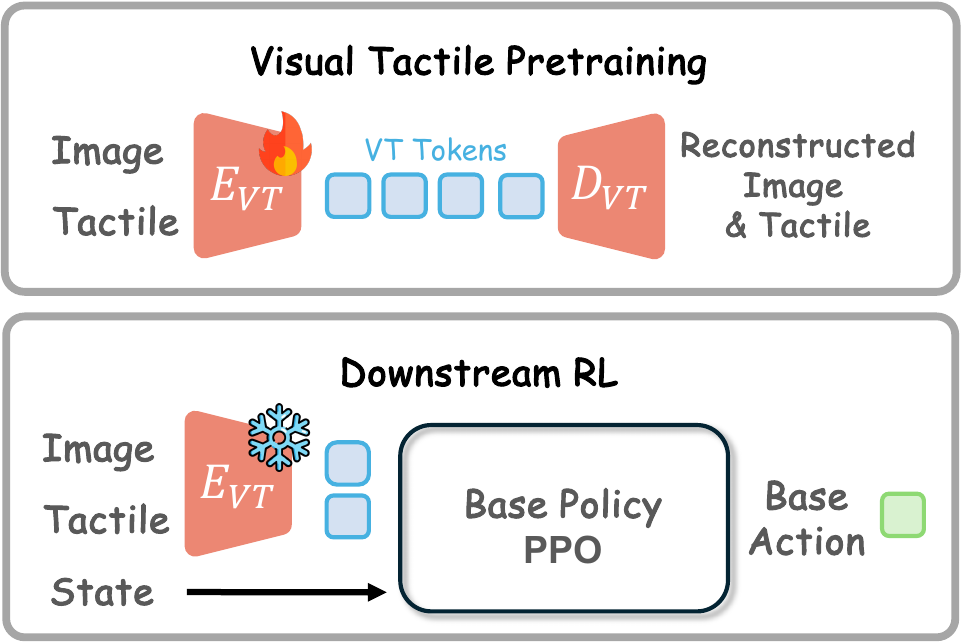}
\caption{VT pretraining pipeline.}
\label{fig:vtpretrain}
\end{figure}

We implement all baseline methods under the same task environments, object
splits, reward functions, and PPO settings. The methods differ only in the
human supervision used during pretraining and whether online guidance is
provided during RL.  
\paragraph{VT pretrain.}
We implement VT following the visual--tactile joint pretraining method of~\cite{liu2025vtdexmanip}. As shown in Fig.~\ref{fig:vtpretrain}, each training sample contains a visual--tactile history,\((V_{t-H+1:t},T_{t-H+1:t})\), with \(H=8\). Each RGB frame is divided into image patches, while each tactile frame is represented by finger level tokens. Tokens from the temporal window are randomly masked, and a transformer encoder fuses the remaining visual and tactile tokens. A lightweight decoder reconstructs the masked inputs.The network and training settings are listed in Table~\ref{tab:vt_fusion_hyperparameters}.After pretraining, the decoder is discarded and the encoder is frozen. The fused visual-tactile feature is concatenated with robot proprioception and used as input to the PPO policy.

\begin{table}[ht]
    \centering
    \small
    \renewcommand{\arraystretch}{1.05}
    \begin{tabular}{lc}
        \toprule
        \textbf{Hyperparameter} & \textbf{All Models} \\
        \midrule
        History length \(H\) & \(8\) \\
        Image resolution per frame \((H_v,W_v)\) & \((224,224)\) \\
        Image patches per frame \(N_v\) & \(196\) \\
        Tactile tokens per frame \(N_c\) & \(5\) \\
        Patch resolution \((P,P)\) & \((16,16)\) \\
        Encoder embedding dimension \(d_{\mathrm{en}}\) & \(384\) \\
        Image mask ratio \(\gamma_v\) & \(0.75\) \\
        Tactile mask ratio \(\gamma_c\) & \(0.50\) \\
        Fusion dimension \(d\) & \(384\) \\
        Decoder embedding dimension \(d_{\mathrm{de}}\) & \(192\) \\
        Image loss weight \(\lambda_v\) & \(1\) \\
        Tactile loss weight \(\lambda_c\) & \(10\) \\
        Learning rate & \(1.5\times10^{-4}\) \\
        Batch size & \(1024\) \\
        \bottomrule
    \end{tabular}
    \caption{Hyperparameters of visual--tactile pretraining.}
    \label{tab:vt_fusion_hyperparameters}
\end{table}

\begin{figure}[ht]
\centering
\includegraphics[width=0.47\textwidth]{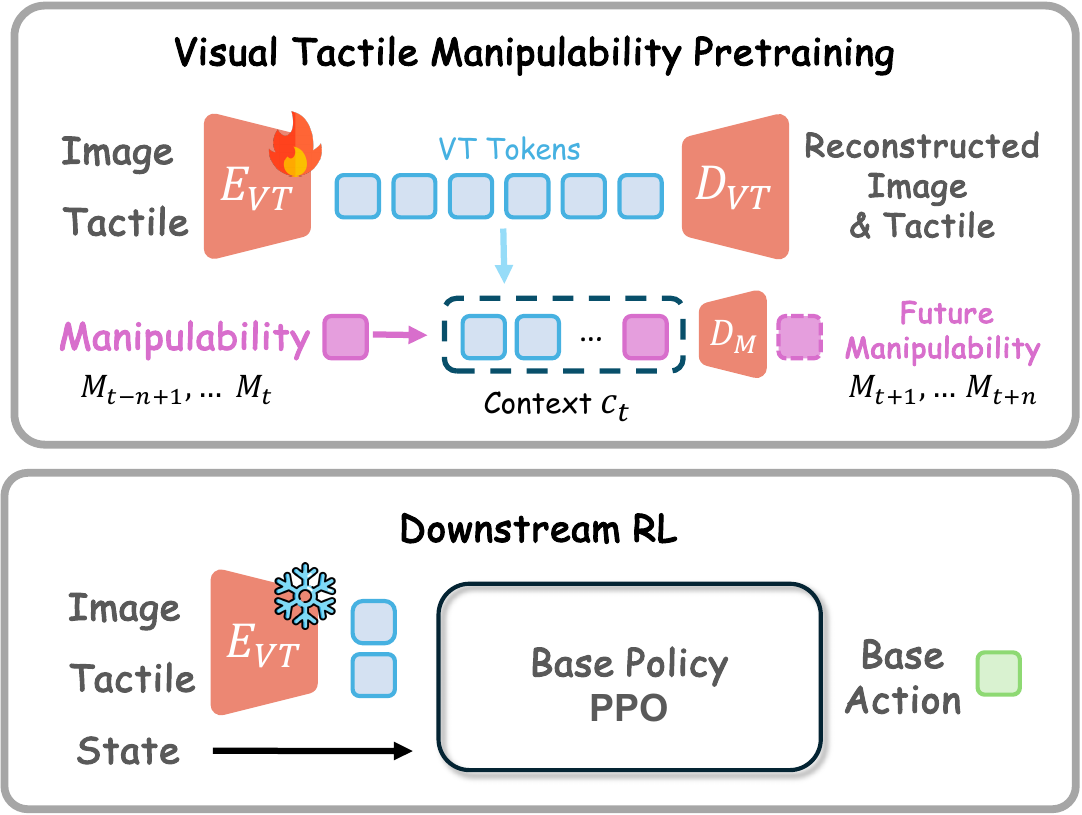}
\caption{VTM pretraining pipeline.}
\label{fig:vtmpretrain}
\end{figure}

\paragraph{VTM.}
As shown in Fig.~\ref{fig:vtmpretrain}, VTM extends VT with manipulability prediction during pretraining. The current contact-conditioned rotational manipulability and task axis are added to the context \(c_t\). predicts four future manipulability descriptors sampled at a stride of four raw frames. The model is trained with manipulability, direction, and magnitude prediction losses. The energy-loss weight is set to
zero, so VTM does not learn an energy field. The detailed settings are listed in Table~\ref{tab:vtm_hyperparameters}.

\begin{table}[ht]
    \centering
    \small
    \renewcommand{\arraystretch}{1.05}
    \begin{tabular}{lc}
        \toprule
        \textbf{Hyperparameters} & \textbf{Value} \\
        \midrule
        Future prediction horizon & \(4\) \\
        Manipulability loss weight & \(1.0\) \\
        Direction loss weight & \(0.2\) \\
        Magnitude loss weight & \(0.1\) \\
        Energy loss weight & \(0.0\) \\
        Batch size & \(1024\) \\
        Learning rate & \(1.5\times10^{-4}\) \\
        \bottomrule
    \end{tabular}
    \caption{Hyperparameters of VTM.}
    \label{tab:vtm_hyperparameters}
\end{table}

\begin{figure}[ht]
\centering
\includegraphics[width=0.47\textwidth]{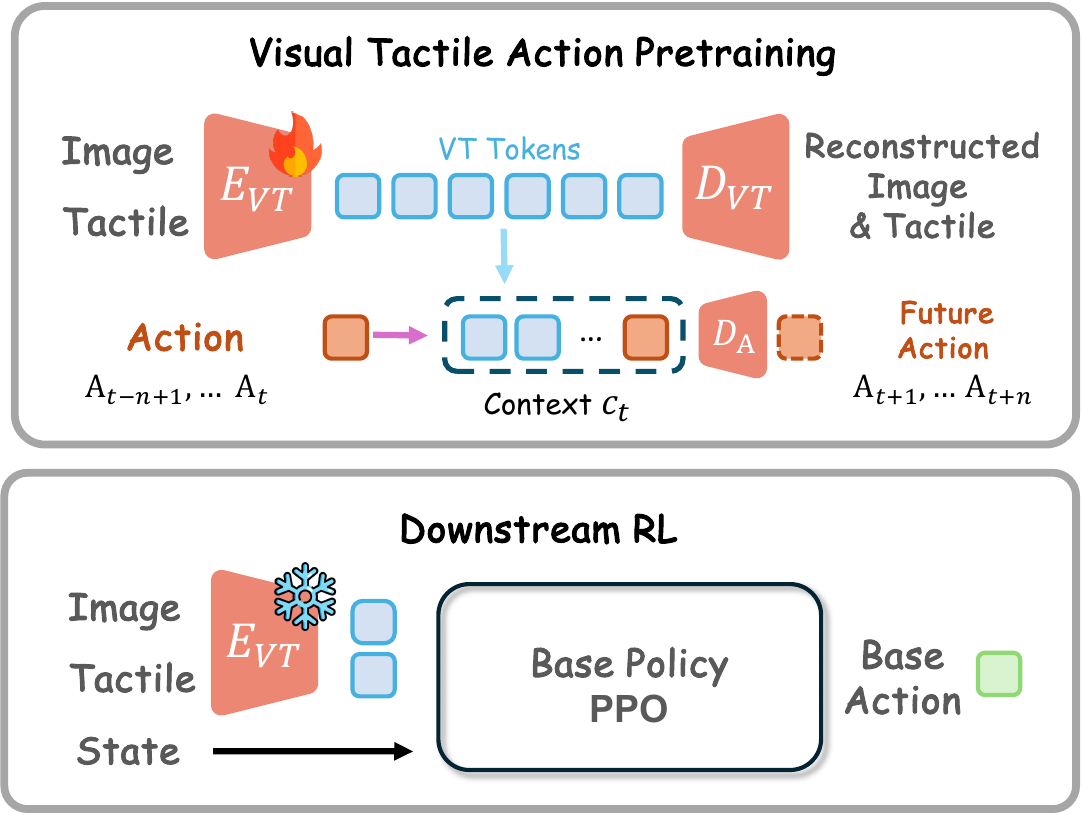}
\caption{VTA pretraining pipeline.}
\label{fig:vtapretrain}
\end{figure}

\paragraph{VTA.}
As shown in Fig.~\ref{fig:vtapretrain}, VTA follows the same pretraining and downstream protocol as VTM. The main difference is the supervision used during pretraining. VTA replaces the manipulability token in the context \(c_t\) with the human action recorded in the dataset. The prediction head also predicts future human actions instead of future manipulability. All other pretraining settings, including the prediction horizon, loss weights, training epochs, batch size, and learning rate, are the same as those of VTM.

\begin{figure}[ht]
\centering
\includegraphics[width=0.47\textwidth]{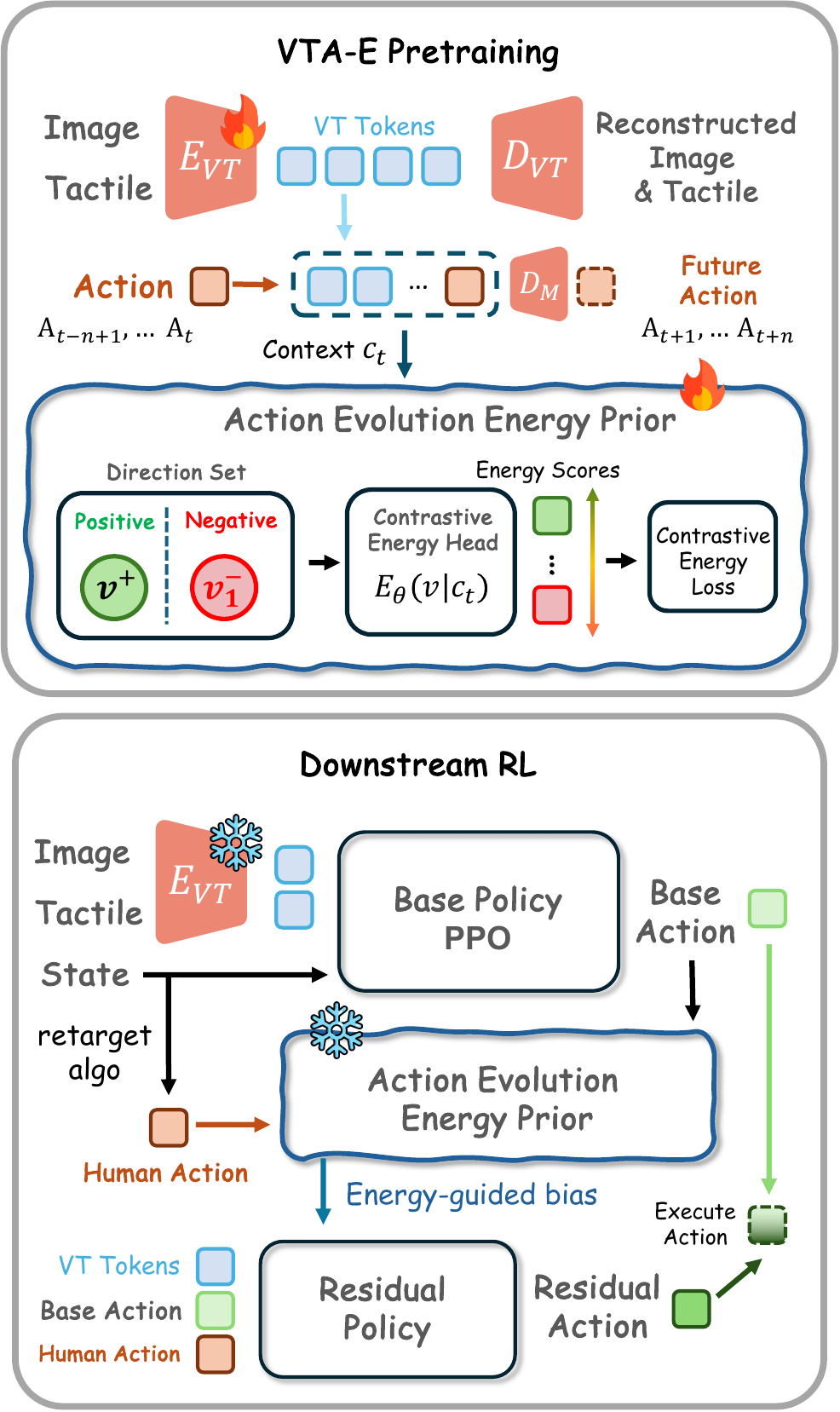}
\caption{VTA-E pretraining pipeline.}
\label{fig:vtaepretrain}
\end{figure}

\paragraph{VTA-E.}
As shown in Fig.~\ref{fig:vtaepretrain},
VTA-E follows the same energy-guided training procedure as VTM-E. The main difference is the representation used by the energy model. During pretraining, VTA-E learns an energy field over human actions. Its auxiliary prediction head also predicts future human actions instead of future manipulability. During downstream RL, both the visual-tactile encoder and the human-action energy model remain frozen. The current robot-hand pose and the poses induced by candidate actions are retargeted to the human-hand space using AnyTeleop~\cite{qin2023anyteleop}. The energy model then scores the corresponding human-action changes, following the same procedure as VTM-E.

\paragraph{Manipulability Evolution Energy Learning.}

During pretraining, manipulability descriptors are sampled every four raw frames. We use the normalized change from \(m_t\) to \(m_{t+4}\) as the positive evolution direction. Each positive sample is paired with eight negatives, including its reversed direction, and randomly sampled directions. Conditioned the context \(c_t\), the energy head assigns a scalar energy to each candidate direction, with lower values indicating better agreement with the human demonstrations. After pretraining, the visual-tactile encoder, context encoder, and energy head are frozen for downstream RL The detailed settings are listed in Table~\ref{tab:energy_pretrain}

\begin{table}[t]
\centering

\small
\setlength{\tabcolsep}{6pt}
\renewcommand{\arraystretch}{1.08}
\begin{tabular}{lc}
\toprule
\textbf{Hyperparameter} & \textbf{Value} \\
\midrule
Temporal offset \(\Delta\) & 4 \\
Prediction horizon & 4 \\
Number of negatives \(K\) & 8 \\
Negative composition & 8 \\
Stationary threshold & \(10^{-4}\) \\
Temperature \(\tau\) & 0.1 \\
Energy loss weight \(\lambda_E\) & 0.1 \\
Magnitude loss weight \(\lambda_\alpha\) & 0.1 \\
Learning rate & \(1.5\times10^{-4}\) \\
Batch size & 1024 \\
Training epochs & 400 \\
\bottomrule
\end{tabular}
\caption{Hyperparameters for energy pretraining.}
\label{tab:energy_pretrain}
\end{table}

\paragraph{PPO Training Details.}
All downstream policies are trained with PPO using the same optimization
settings. We run 200 parallel environments and each policy update
uses rollouts of 32 steps. The collected samples are divided into four
mini-batches and optimized for ten epochs.

The actor and critic use multilayer perceptrons with hidden dimensions
\([1024,1024,512]\). ELU is used as the activation function. The learning rate
is \(3\times10^{-4}\). The PPO clipping range is \(0.2\), and the gradient norm
is clipped at \(1.0\). We use a discount factor of \(\gamma=0.96\) and a GAE
parameter of \(\lambda_{\mathrm{GAE}}=0.95\). The initial policy noise standard
deviation is \(0.8\). The desired KL divergence is \(0.016\), and no entropy
bonus is applied.

The maximum episode length is 500 steps for Unscrew Cap and Turn Faucet, and
600 steps for Rotate Object. The same PPO settings are used for all methods.
The pretrained encoders and energy models remain frozen during downstream
policy training.

\begin{table}[ht]
    \centering
    
    \label{tab:ppo_hyperparameters}
    \small
    \setlength{\tabcolsep}{5pt}
    \renewcommand{\arraystretch}{1.08}
    \begin{tabular}{lc}
        \toprule
        \textbf{Hyperparameter} & \textbf{Value} \\
        \midrule
        Parallel environments & 200 \\
        Rollout length & 32 \\
        Number of mini-batches & 4 \\
        Optimization epochs & 10 \\
        Hidden dimensions & \([1024,1024,512]\) \\
        Activation & ELU \\
        Learning rate & \(3\times10^{-4}\) \\
        PPO clip range & \(0.2\) \\
        Maximum gradient norm & \(1.0\) \\
        Discount factor \(\gamma\) & \(0.96\) \\
        GAE parameter \(\lambda_{\mathrm{GAE}}\) & \(0.95\) \\
        Initial action-noise std. & \(0.8\) \\
        Desired KL divergence & \(0.016\) \\
        Entropy coefficient & \(0\) \\
        \bottomrule
    \end{tabular}
    \caption{PPO training hyperparameters.}
\end{table}

\paragraph{Domain Randomization.}
\label{app:sim2real}

We apply domain randomization~\cite{tobin2017domainrandomizationtransferringdeep} during simulation training to reduce the gap between simulated and real observations. The randomization covers proprioception, appearance, camera configuration, and tactile sensing.

Zero-mean Gaussian noise is added to the joint angles and joint velocities at each observation step. Object color and light direction are randomized when the environment is reset. We also perturb the camera position and look-at point at each reset. Image brightness, contrast, and pixel values are randomized independently for every frame. Finally, Gaussian noise is added to the tactile force readings before contact binarization. The complete settings are listed in Table~\ref{tab:sim2real_randomization}.

\begin{table}[ht]
    \centering
    \small
    \renewcommand{\arraystretch}{1.08}
    \begin{tabular}{@{}ll@{}}
        \toprule
        \textbf{Parameter} & \textbf{Randomization strategy} \\
        \midrule
        Joint angle
            & \(+\mathcal N(0,0.05)\), per observation \\
        Joint velocity
            & \(+\mathcal N(0,0.05)\), per observation \\
        \midrule
        Object color
            & Randomized at environment reset \\
        Light direction
            & Randomized at environment reset \\
        Camera eye position
            & \(\mathcal U[-4\,\mathrm{cm},4\,\mathrm{cm}]\), at reset \\
        Camera look-at position
            & \(\mathcal U[-4\,\mathrm{cm},4\,\mathrm{cm}]\), at reset \\
        Image brightness
            & \(\mathcal U[-20,20]\), per frame \\
        Image contrast
            & \(\mathcal U[-8,8]\), per frame \\
        Image pixel noise
            & \(\mathcal U[-5,5]\), per frame \\
        \midrule
        Tactile force
            & \(+\mathcal N(0,0.1)\), before binarization \\
        \bottomrule
    \end{tabular}
    \caption{\textbf{Domain randomization settings.}
    \(\mathcal N\) and \(\mathcal U\) denote Gaussian and uniform
    distributions, respectively.}
    \label{tab:sim2real_randomization}
\end{table}

At deployment, the real visual, tactile, and proprioceptive observations are
processed using the same input format as in simulation. The tactile signals
are binarized before being passed to the policy. No real-world policy
fine-tuning is performed.

\section{4 Experiment Details}

DexMani uses human demonstrations to pretrain the visual--tactile
encoder and the prior over contact-conditioned rotational manipulability
evolution.  
\subsection{4.1 Task Details}

\begin{figure}[ht]
\centering
\includegraphics[width=0.47\textwidth]{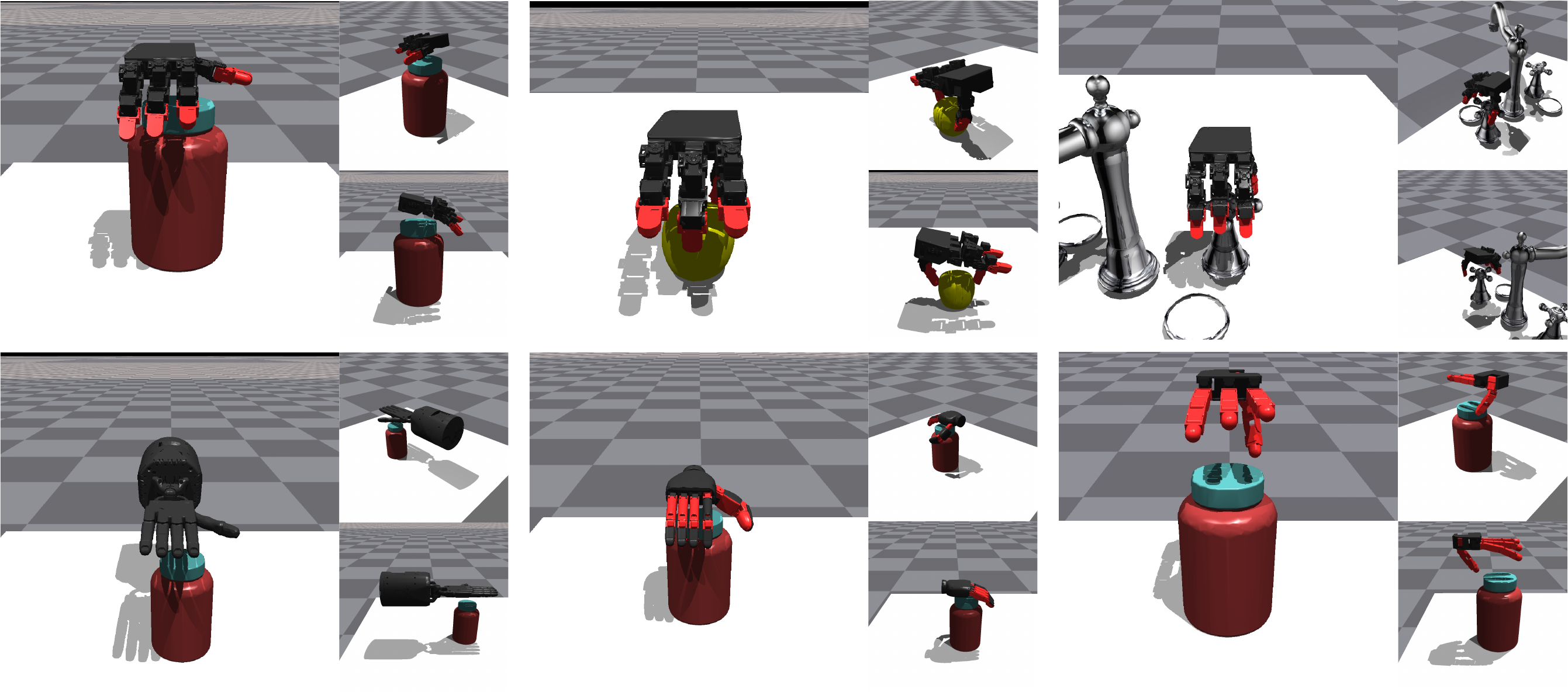}
\caption{\textbf{Simulated task settings.}
Top: Unscrew Cap, Rotate Object, and Turn Faucet on the LEAP Hand.
Bottom: cap unscrewing with the Shadow, XHand, and Allegro Hand.}
\label{fig:leapsettings}
\end{figure}

\begin{figure}[ht]
\centering
\includegraphics[width=0.47\textwidth]{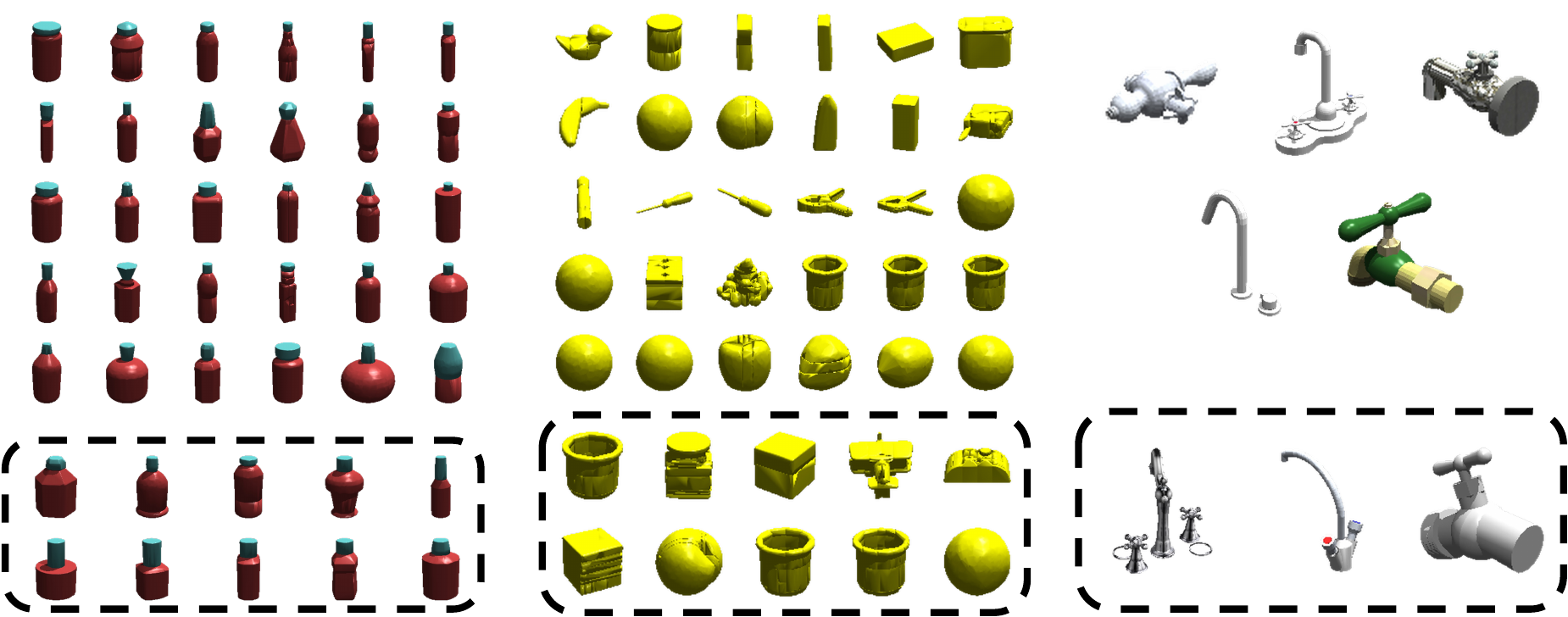}
\caption{\textbf{Objects used in the three simulation tasks.}
Objects inside the black dashed boxes are held out for unseen-object
evaluation.}
\label{fig:simobjs}
\end{figure}

We evaluate DexMani on three contact-rich object rotation tasks: \textit{Unscrew Cap}, \textit{Rotate Object}, and \textit{Turn Faucet}. These tasks require different contact patterns and finger coordination strategies, while sharing the goal of sustained object rotation.
Figure~\ref{fig:leapsettings} shows the three LEAP Hand environments, and the setup used for cross-hand evaluation on the Shadow Hand, Allegro Hand, and XHand in IsaacGym~\cite{makoviychuk2021isaacgymhighperformance}. Figure~\ref{fig:simobjs} shows the training and evaluation objects used in the simulation tasks. Objects inside the black dashed boxes are excluded from policy training and used only for unseen-object evaluation. The simulation environments and reward functions are adapted from VTDexManip~\cite{liu2025vtdexmanip}. We adjust the task settings to match our evaluation protocol. The setup, success criterion, object split, and reward function of each task are described below.

\paragraph{Unscrew Cap.}

This task evaluates coordinated finger motion during bottle-cap rotation.
Each bottle is fixed to the table. The cap has one rotational degree of
freedom about its local \(z\)-axis. The goal is to rotate the cap
counterclockwise by one full revolution within an episode.

The reward is
\begin{equation}
r
=
\lambda_1 r_p
+
\lambda_2 r_v
+
\lambda_3 r_d
+
\lambda_4 r_s .
\end{equation}
The rotation term is
\(r_p=\min(\theta_{\mathrm{joint}},7.0)\), where \(\theta_{\mathrm{joint}}\) is the cap rotation angle. The velocity term is \(r_v=\operatorname{clamp}(v_{\mathrm{joint}},-10,10)\). The distance term is \(r_d=\exp(-10d)\), where \(d\) is the summed distance from the fingertips to a reference point located \(2\,\mathrm{cm}\) below the cap. The success term is \(r_s=5\). The reward weights are listed in Table~\ref{tab:unscrew_reward}.

\begin{table}[ht]
\centering
\small
\begin{tabular}{lc}
\toprule
\textbf{Reward term} & \textbf{Weight} \\
\midrule
Rotation position & \(\lambda_1=0.5\) \\
Angular velocity & \(\lambda_2=1\) \\
Fingertip distance & \(\lambda_3=0.5\) \\
Success & \(\lambda_4=1\) \\
\bottomrule
\end{tabular}
\caption{Reward weights for Unscrew Cap.}
\label{tab:unscrew_reward}
\end{table}

The velocity reward is disabled when the cap has positive angular velocity
but no tactile contact is active. This encourages the hand to rotate the cap
through fingertip contact.

\paragraph{Rotate Object.}
In this task, the dexterous hand rotates an object on a table without toppling it. The task requires coordination between the thumb and the other fingers. The goal is to rotate the object by \(2\pi\) while keeping it upright. We use thirty YCB objects~\cite{Calli_2015} for policy training and ten held-out objects for evaluation.

The reward is
\begin{equation}
r
=
{\lambda}_1^\top r_d
+
\lambda_2 r_{\mathrm{rot}}
+
\lambda_3 r_a
+
\lambda_4 r_v
+
r_b .
\end{equation}
The distance term is
\(r_d=[\exp(-10d_1),\,d_2]^\top\).
Here, \(d_1\) is the vertical distance between the fingertips and the object. The term \(d_2\) is the planar distance between the object and its target position. The term \(r_{\mathrm{rot}}\) measures the difference between the current and target object poses. The action penalty is \(r_a=\lVert a\rVert_2\). The velocity term is \(r_v=\operatorname{clamp}(\omega_z,-10,10)\), where \(\omega_z\) is the object angular velocity about the \(z\)-axis. A success bonus of \(r_b=250\) is added when the task is completed. The reward weights are listed in Table~\ref{tab:rotate_reward}.

\begin{table}[ht]
\centering
\small
\begin{tabular}{lc}
\toprule
\textbf{Reward term} & \textbf{Weight} \\
\midrule
Distance
& \(\boldsymbol{\lambda}_1=[0.25,-10]^\top\) \\

Pose difference
& \(\lambda_2=1\) \\

Action penalty
& \(\lambda_3=-2\times10^{-4}\) \\

Angular velocity
& \(\lambda_4=1\) \\

Success bonus
& \(1\) \\
\bottomrule
\end{tabular}
\caption{Reward weights for Rotate Object.}
\label{tab:rotate_reward}
\end{table}

\paragraph{Turn Faucet.}
This task requires the dexterous hand to rotate a faucet handle. Unlike Unscrew Cap, the fingers must push against the handle rather than twist a cylindrical cap. The goal is to rotate the handle clockwise by one full revolution. We use five faucet models from the SAPIEN dataset~\cite{Xiang_2020_SAPIEN} for policy training. All models have a rotation axis perpendicular to the ground.
For evaluation, we use three unseen instances. The reward for faucet turning follows the same formulation as Unscrewing cap. The main difference is that the velocity reward is always applied and is not gated by tactile contact. In addition, we encourage the fingertips to approach task-specific target positions around the faucet handle. The distance term is computed as the sum of the distances from each fingertip to its corresponding target position.

\subsection{4.2 Additional Analysis}

\begin{figure}[t]
\centering
\includegraphics[width=0.47\textwidth]{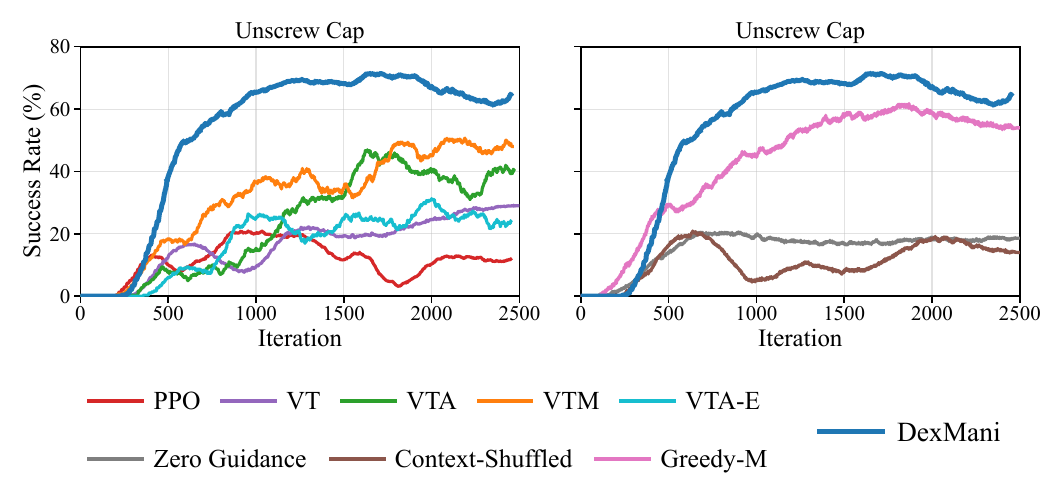}
\caption{Learning curves of unscrew cap Task.}
\label{fig:learningcurves}
\end{figure}

\begin{figure}[t]
\centering
\includegraphics[width=0.47\textwidth]{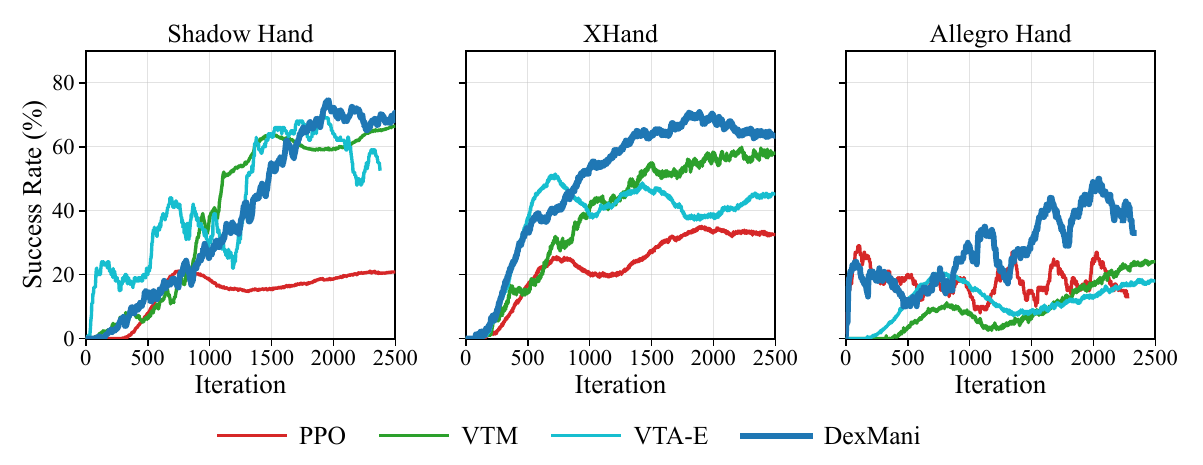}
\caption{Learning curves of cross-hand unscrew cap Task.}
\label{fig:crosshand}
\end{figure}

\paragraph{Learning efficiency.} Figures~\ref{fig:learningcurves} and~\ref{fig:crosshand} compare learning progress in simulation. On the LEAP Hand, DexMani reaches strong cap-unscrewing performance earlier and converges to a higher final success rate than the baselines. The same trend holds across the Shadow, Allegro, and XHand embodiments, despite their different kinematics and action spaces. These results show that the human-derived guidance improves both learning efficiency and final policy performance, and that this benefit remains consistent across hand morphologies.

\begin{figure}[t]
\centering
\includegraphics[width=0.47\textwidth]{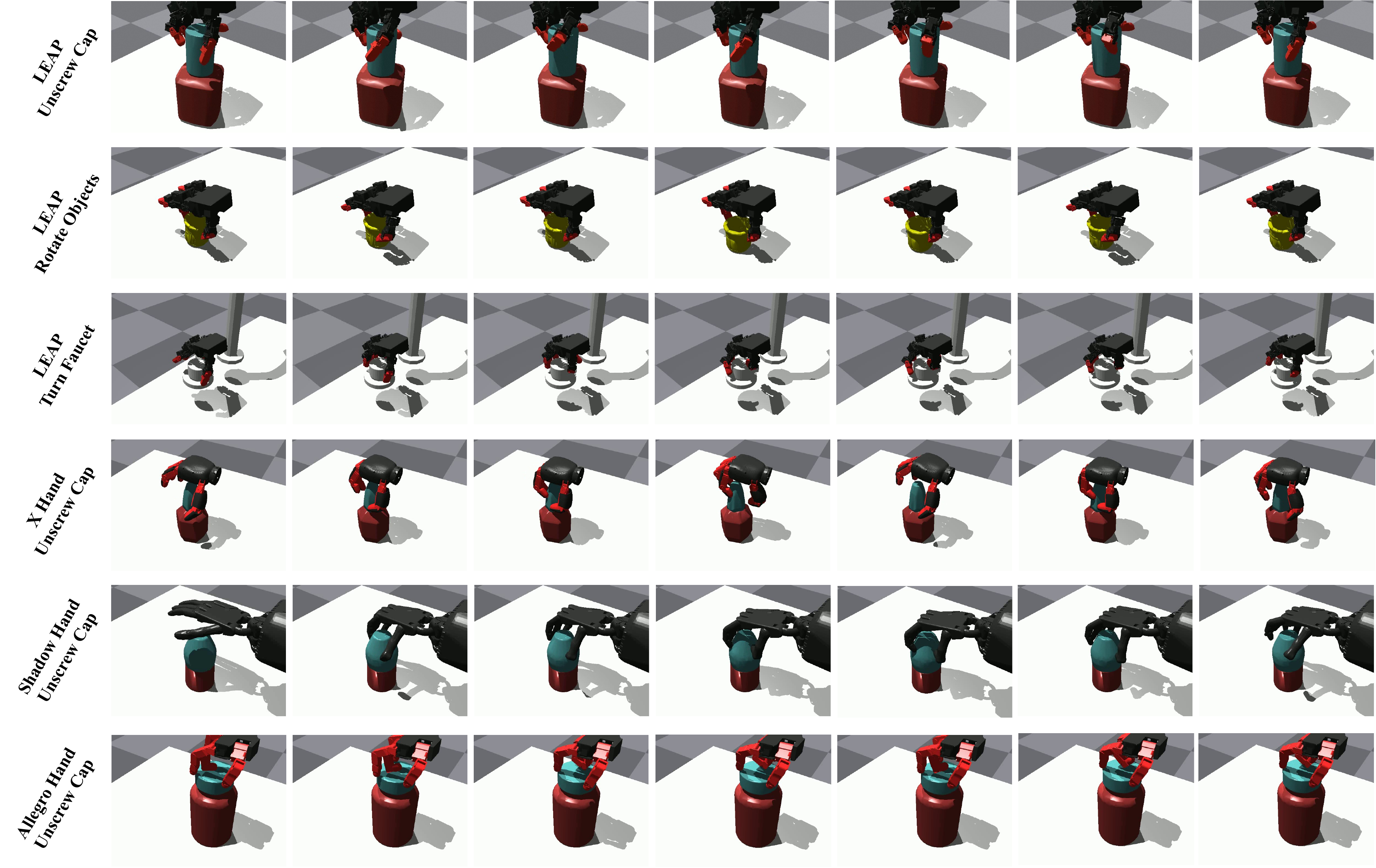}
\caption{\textbf{Representative simulated rollouts.}
Six sequences illustrate sustained rotation through repeated contact transitions across the evaluated settings.}
\label{fig:simsequences}
\end{figure}

Figure~\ref{fig:simsequences} further shows six representative simulated rollouts. Across different settings, DexMani maintains object progress while releasing and re-establishing finger contacts, producing sustained rotation without prescribing a fixed contact sequence. Together with the learning curves, these examples illustrate that the human-derived guidance improves both policy learning and execution.

\paragraph{Real-world experiment.} The Faucet task presents the largest distribution shift in our real-world evaluation. Neither faucet-turning demonstrations nor the physical faucet object appear in the human pretraining data, and the physical faucet also differs from the simulated objects used for robot-policy training. The policy must therefore handle a new task-specific contact pattern, unseen object geometry, and the sim-to-real sensing and dynamics gap simultaneously. We attribute the low success rate mainly to this combined task and object distribution shift.